\documentclass[runningheads]{llncs}

\usepackage[mobile]{eccv}

\usepackage{times}
\usepackage{epsfig}
\usepackage{graphicx}
\usepackage{amsmath}
\usepackage{amssymb}
\usepackage{booktabs}
\usepackage{lipsum}
\usepackage{mwe}
\usepackage{multirow}
\usepackage{rotating}
\usepackage{adjustbox}
\usepackage{tabularx}
\usepackage{array}

\usepackage{eccvabbrv}

\usepackage{graphicx}
\usepackage{booktabs}

\usepackage[accsupp]{axessibility}  %

\usepackage[pagebackref,breaklinks,colorlinks,citecolor=eccvblue]{hyperref}
\usepackage{orcidlink}

\newcommand{\ourM}{VLAD-BuFF}

\begin{document}

\title{{VLAD-BuFF: Burst-aware Fast Feature Aggregation for Visual Place Recognition}} 

\titlerunning{\ourM{}}

\author{Ahmad Khaliq\inst{1}\orcidlink{0000-0002-7878-9369} \and
Ming Xu\inst{2}\orcidlink{0000-0002-6478-0582} \and
Stephen Hausler\inst{3}\orcidlink{0000-0003-0092-0096} \and \\
Michael Milford\inst{1}\orcidlink{0000-0002-5162-1793} \and
Sourav Garg\inst{4}\orcidlink{0000-0001-6068-3307}}

\authorrunning{A. Khaliq et al.}

\institute{Queensland University of Technology, Australia \and
Australian National University, Australia \and
CSIRO, Australia \and
University of Adelaide, Australia}

\maketitle

\begin{abstract}
Visual Place Recognition (VPR) is a crucial component of many visual localization pipelines for embodied agents. VPR is often formulated as an image retrieval task aimed at jointly learning local features and an aggregation method. The current state-of-the-art VPR methods rely on VLAD aggregation, which can be trained to learn a weighted contribution of features through their soft assignment to cluster centers. However, this process has two key limitations. Firstly, the feature-to-cluster weighting does not account for over-represented repetitive structures \textit{within} a cluster, e.g., shadows or window panes; this phenomenon is also referred to as the `burstiness' problem, classically solved by discounting repetitive features before aggregation. Secondly, feature to cluster comparisons are compute-intensive for state-of-the-art image encoders with high-dimensional local features. This paper addresses these limitations by introducing VLAD-BuFF with two novel contributions: i) a self-similarity based feature discounting mechanism to learn \textit{\textbf{Bu}rst-aware} features within end-to-end VPR training, and ii) \textit{\textbf{F}ast \textbf{F}eature aggregation} by reducing local feature dimensions \textit{specifically} through PCA-initialized learnable pre-projection. We benchmark our method on $9$ public datasets, where VLAD-BuFF sets a new state of the art. Our method is able to maintain its high recall even for $12\times$ reduced local feature dimensions, thus enabling fast feature aggregation without compromising on recall. Through additional qualitative studies, we show how our proposed weighting method effectively downweights the non-distinctive features. Source code: \url{https://github.com/Ahmedest61/VLAD-BuFF/}.

\end{abstract}

\section{Introduction}
\label{sec:intro}

Visual Place Recognition (VPR)~\cite{ zhu2018attention, toriivprrepet, camara2019spatio, hong2019TextPlace, chen2017only, cummins2008fab, nowicki2017real,garg2021where} is crucial for autonomous embodied agents to localize, navigate and function in the world, e.g., driverless cars, drones and augmented/virtual reality devices. Image retrieval based VPR methods are typically based on compact vector representations of images, enabling fast and coarse visual localization as a precursor to precise 3D pose estimation. These global image representations are based on aggregation of local features, for example, Bag of Words (BoW)~\cite{sivic2003video} and Vector of Locally Aggregated Descriptors (VLAD)~\cite{jegou2010aggregating}. In the last decade, \textit{deep learning-based} image representations have been the primary focus of retrieval-based VPR. This includes innovations on several fronts: backbone networks~\cite{keetha2023anyloc,berton2022deep,wang2022transvpr}, objective functions~\cite{berton2023eigenplaces,revaud2019learning,yu2019spatial,thoma2020soft}, multi-scale processing~\cite{xin2019real,yu2019spatial,hausler2021patchnetvlad}, semantics~\cite{gawel2018x,garg2018lost,paolicelli2022learning,garg2024robohop}, sequences~\cite{ho2007detecting,milford2012seqslam,neubert2019neurologically,schubert2021fast,garg2021seqnet}, as well as aggregation/pooling~\cite{izquierdo2023optimal,arandjelovic2016netvlad,radenovic2018fine,chen2021learning,tolias2015particular,zhou2016learning,wang2022transvpr}. Learning-based pooling methods aggregate local features through exponentiation (e.g., GAP~\cite{zhou2016learning,oertel2020augmenting}, MAC~\cite{tolias2015particular}, and GeM~\cite{radenovic2018fine}) or weighting (e.g., NetVLAD~\cite{arandjelovic2016netvlad}, SALAD~\cite{izquierdo2023optimal}, Context Gating~\cite{miech2017learnable}, Attention~\cite{wang2022transvpr,cao2020unifying}, and GPO~\cite{chen2021learning}). The current state-of-the-art methods in VPR are based on VLAD or its variants, e.g., AnyLoc~\cite{keetha2023anyloc} and SALAD~\cite{izquierdo2023optimal}. High accuracy of these methods is attributed to their specific design aspects of how clusters centers are created, how features-to-cluster assignment is performed, and whether or not residuals are computed. 
Complementary to these, in this paper, we consider two additional aspects of deep-learnt VLAD aggregation to further improve its effectiveness and practical utility: \textit{a)} feature-to-feature (self-similarity) based weighting to deal with feature burstiness problem~\cite{jegou2009burstiness} and \textit{b)} feature dimension reduction \textit{before} aggregation to enable faster aggregation without compromising on recall.   

NetVLAD~\cite{arandjelovic2016netvlad} and its follow-up VLAD enhancements in VPR~\cite{miech2017learnable,hausler2021patchnetvlad,yu2019spatial,khaliq2022multires,le2020city,mereu2022learning,keetha2021hierarchical,keetha2023anyloc,izquierdo2023optimal}, including the most recent work SALAD~\cite{izquierdo2023optimal}, are all reliant on only the feature-to-cluster weighting. Such weighting does not account for over-representation of repetitive features, e.g., shadows, clouds, road, and window panes. This can inadvertently under-represent other relatively salient elements within a cluster, e.g., unique features from signboards or specific features from buildings and vegetation. This issue is also referred to as the `feature burstiness' problem~\cite{jegou2009burstiness}, where the classical solution is to discount repetitive features by \textit{post-processing} self-similarity of \textit{non-learnable} features~\cite{jegou2009burstiness,toriivprrepet,delhumeau2013revisiting}. However, in the context of modern deep learning-based methods, these solutions need to be reconsidered.
In this vein, for the first time, we incorporate `burstiness-awareness' within end-to-end VPR training to jointly learn the features under the influence of a learnable burstiness-discounting mechanism. Specifically, we propose a novel \textit{soft counting} layer based on feature-to-feature similarities, which is used to discount the vanilla feature-to-cluster soft assignment weights. 

The high accuracy of VLAD comes at the cost of high compute time due to its cluster-wise aggregation process. This becomes a significant bottleneck for powerful image encoder backbones which typically output high-dimensional features. The obvious solution to this is to simultaneously learn a linear/non-linear projection to a low-dimensional space. However, such \textit{randomly-initialized} projection adversely affects the initialization of cluster centers and feature-to-cluster assignment. We show that using \textit{PCA-based} initialization for such feature projection is much more effective and retains superior recall even at significantly lower dimensions, thus substantially reducing the aggregation time.

On a diverse range of standard benchmark datasets, we show that \ourM{} sets a new state of the art, while \textit{consistently} outperforming previous approaches, including the very recent methods such as SALAD~\cite{izquierdo2023optimal} and AnyLoc~\cite{keetha2023anyloc}. Furthermore, our proposed pre-pool PCA projection unveils a new efficiency-inducing paradigm for learning-based VLAD aggregation, which maintains superior recall even for $12\times$ smaller local descriptor dimensions. 
Finally, we provide insights into how visual burstiness phenomenon affects retrieval performance in the context of learned VPR methods through ablation studies and qualitative examples.

\section{Related Works}
\subsection{Visual Place Recognition}
In this section, we provide a brief overview of learning-based VPR. Detailed literature review on the VPR task more generally is provided in~\cite{lowry2016visual, masone2021survey, garg2021where, zhangsurvey, tsintotas2022revisiting, yin2022general, schubert2023visual}. An important insight into the VPR task is that it can be formulated as a retrieval task~\cite{garg2021where}. As a result, many of the recent developments in learning-based VPR revolve around developing novel retrieval-based loss functions. For example, NetVLAD~\cite{arandjelovic2016netvlad} uses a triplet margin loss with hard-negative mining, AP-GeM~\cite{revaud2019learning} uses a listwise loss, CosPlace~\cite{berton2022rethinking} uses a classification style loss,~\cite{thoma2020soft} adjusts triplet loss to have soft positives/negatives, reflecting continuous GPS data, EigenPlaces~\cite{berton2023eigenplaces} uses viewpoint variation based classification loss, and MixVPR~\cite{ali2023mixvpr} uses multi-similarity loss~\cite{wang2019multi} based on self- and relative-similarity weighting of sample pairs. In this work, we consider a variety of training losses in conjunction with the proposed aggregation method to demonstrate complementary performance gains.

There have been advancements around identifying suitable network architectures for the VPR task. Network architectures for VPR can be broadly partitioned into a feature extraction backbone and pooling module. Both CNN~\cite{arandjelovic2016netvlad, revaud2019learning} and ViT-based~\cite{keetha2023anyloc,izquierdo2023optimal,dosovitskiy2021an, wang2022transvpr} architectures have been used for the feature extraction backbone, and we will discuss pooling layers in the following paragraph. Recently, the VG benchmark~\cite{berton2022deep} performed a comprehensive comparison between a number of popular loss functions, feature extractor backbone architectures and pooling layers for the VPR task. Our method in particular, modifies the pooling/aggregation layer. Therefore, it is complementary to advances in loss functions, training dataset choice, and backbone architectures, as demonstrated in our experiments and ablation studies.

\subsection{Pooling/Aggregation Types}
VPR methods generally come with a pooling or aggregation layer, which compresses the spatial axes of the local feature tensor into a single $D$-dimensional embedding called a \textit{global descriptor}. The pooling layer is a heavily studied topic in the retrieval and VPR literature, for example, MAC~\cite{tolias2015particular}, GAP~\cite{zhou2016learning, oertel2020augmenting}, GeM~\cite{radenovic2018fine,revaud2019learning}, VLAD~\cite{arandjelovic2016netvlad} and those based on attention~\cite{wang2022transvpr}, sorting~\cite{chen2021learning} and flattening~\cite{chen2014convolutional,ali2023mixvpr}. Much recent works, such as AnyLoc~\cite{keetha2023anyloc} and SALAD~\cite{izquierdo2023optimal}, have demonstrated state-of-the-art results in VPR using variants of VLAD aggregation. In this work, we improve on the VLAD aggregation layer, originally proposed in NetVLAD~\cite{arandjelovic2013all}, by accounting for the visual burstiness phenomena (explained next) and local feature projection \textit{before} aggregation. We show that addressing these aspects respectively leads to better accuracy and aggregation speed. Furthermore, both our contributions remain complementary to prior innovations in learning-based VLAD design, which includes variations in choices on use of multiple scales~\cite{hausler2021patchnetvlad, yu2019spatial,khaliq2022multires,le2020city}, sequences~\cite{mereu2022learning}, cluster selection~\cite{keetha2021hierarchical}, cluster initialization~\cite{izquierdo2023optimal}, vocabulary selection~\cite{keetha2023anyloc}, residual exclusion~\cite{miech2017learnable,izquierdo2023optimal} and assignment strategy~\cite{izquierdo2023optimal,keetha2023anyloc}. 

\subsection{Visual Burstiness}

The burstiness phenomenon was initially observed in text retrieval and addressed using the well-known tf-idf statistic~\cite{saltontext}. Tf-idf has also been used in image retrieval and VPR using bag-of-words (BoW) image representations~\cite{sivic2003video,cummins2008fab}. A more concrete definition of visual burstiness was proposed in~\cite{jegou2009burstiness}, based on the violation of statistical independence of visual word occurrences.~\cite{jegou2009burstiness} also provides a simple method to improve retrieval performance by reducing the effect of burstiness.~\cite{toriivprrepet} extended~\cite{jegou2009burstiness} by defining burstiness to occur within spatially localised groups. Following this, \cite{manandharbursty} proposes a retrieval metric, named FRS, for deep learnt features, which also accounts for bursty features in a geometrically consistent way.~\cite{guissous19} used DBSCAN to cluster local features into groups to identify bursty elements before aggregation, whereas~\cite{shi2015early} used several explicit burst detection strategies to then perform asymmetric aggregation and matching.~\cite{trichetbursty19} proposed to account for burstiness through Gaussian normalization to preserve distribution information. In the context of VLAD aggregation,~\cite{arandjelovic2013all} proposed intra-normalization to balance burstiness across cluster centers and~\cite{delhumeau2013revisiting} demonstrated that the combined use of residual normalization and power law normalization can better handle bursts caused by features closer to their cluster centers. All aforementioned techniques seek to identify and downweight the contribution of bursty local features to the final global pooled representation. 
Unlike these methods, where adjustments for burstiness are mostly independent of the backend features, we include the burstiness-discounted weighting directly within the end-to-end training for the VPR task. We demonstrate a consistent recall improvement across a wide selection of benchmark datasets through the proposed joint learning of burstiness weights and visual features.

\subsection{Feature Weighting} More generally, our proposed method falls into the category of learned local feature weighting before pooling, of which there are existing works. For example,~\cite{zhu2018attention} incorporates a learned attention block to weight feature residuals for VPR and~\cite{linnextvlad18, kmiec2018learnable} apply attention to VLAD residuals for video classification. For point cloud data,~\cite{zhangpcan19, xia2021soe, sun2020dagc, pengattention21} used learned soft attention to weight individual point features before VLAD aggregation.~\cite{zhu2018attention} uses a cascaded attention block for feature pyramid aggregation. In contrast to these approaches, our method specifically aims to counteract the visual burstiness phenomenon to enhance the distinctiveness and robustness of the feature representation.

\begin{figure}[t]
    \includegraphics[width=\textwidth]{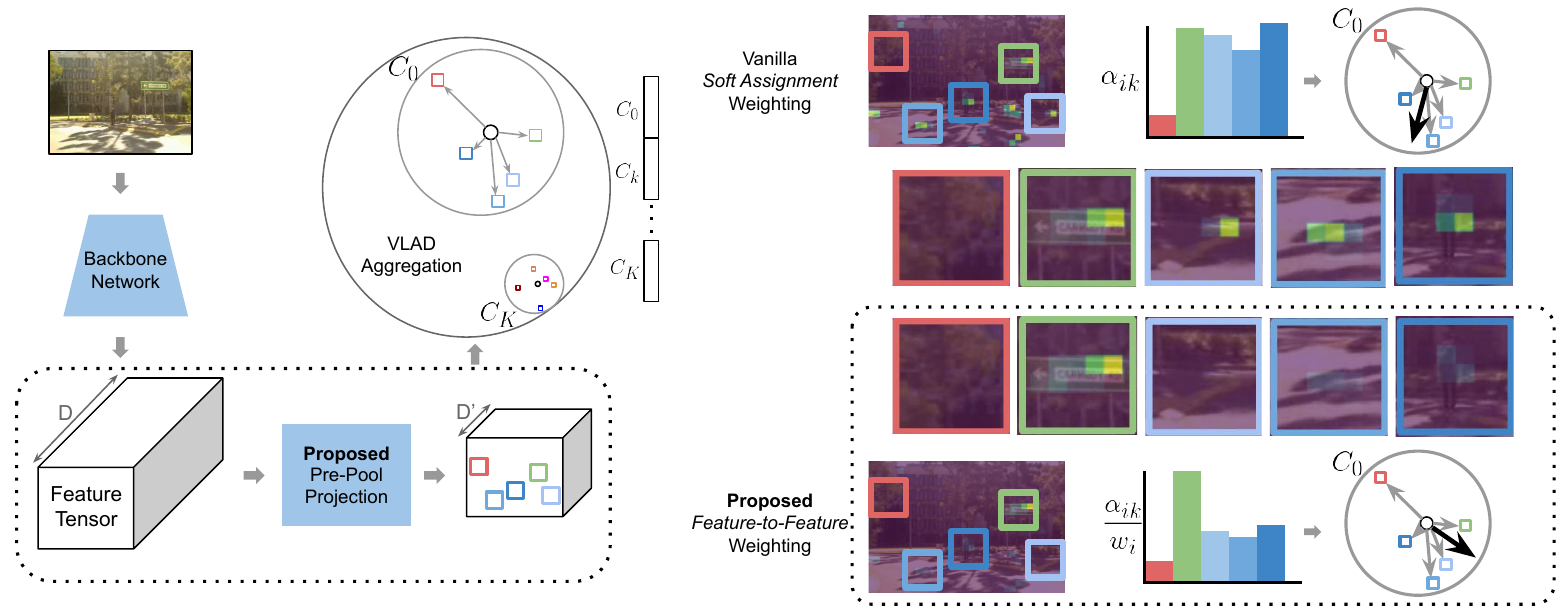}
    \caption{An illustration of our proposed method. \textit{Left:} within a typical VLAD aggregation pipeline, we propose a pre-pool PCA projection layer (dotted box) to reduce the local feature dimensions (from $D$ to $D'$), enabling a compute-efficient aggregation. For a given cluster ($C_k$) on a unit hypersphere, residual vectors (gray arrows) are shown between centroids (tiny circles) and local features (colored squares). \textit{Right:} for a set of local features assigned to cluster $C_0$, we show their weighting pattern through histograms (color corresponds to local features) as well as zoomed image patches with overlaid masks (yellow represents a high weight value). For vanilla NetVLAD's weighting based solely on soft assignment, several features on shadows (blue in histogram) cause a burst and are weighted highly, as opposed to our proposed method which balances out the weights on the burst but has a high weight value for features on the signboard (green in histogram). This variation in the weighting pattern changes the relative contribution of local features to the aggregated vector. This can be observed from the change in the direction of the aggregated vector shown as a black arrow in $C_0$ at the far right; our method moves the aggregated vector closer to the highly-weighted signboard feature.}
    \label{fig:pipeline}
\end{figure}

\section{VLAD-BuFF}
We aim to learn a compact vector representation $V$ of a given image $I$ for the task of visual place recognition, as depicted in \cref{fig:pipeline}. We base this representation on NetVLAD~\cite{arandjelovic2016netvlad}, which learns to aggregate the weighted cluster residuals of $N$ $D$-dimensional local descriptors $\mathbf{x}_i$ for each of the $C$ cluster centers $\mathbf{c}_k$. We propose to `buff' this representation with a learnable burstiness weighting $w_i$ based on self-similarity $d_{ij}$ of local descriptors within an image. In particular, before aggregation, we intend to weight the contribution of a local descriptor to its residual aggregation by the \textit{count} of descriptors that are `similar' to it, which is achieved by measuring feature-to-feature distances. This is applied on top of vanilla NetVLAD's feature weighting, which is solely based on $1\times1$ convolutional soft assignment initialized through feature-to-centroid distances. While the previous approaches to burstiness weighting were based on hand-crafted features~\cite{jegou2009burstiness,toriivprrepet,delhumeau2013revisiting}, we aim to learn the local features and their corresponding weights end-to-end using a place recognition training objective.

\subsection{Soft Count} A count of similar items in a set can be defined as the sum of binary decisions about whether or not an element is similar to others (and itself). For elements which lie in a continuous vector space, distance thresholds (typically based on Euclidean distance) are used to define whether or not two elements are considered similar or not. These binary decisions therefore become a non-differentiable function at the decision threshold. We propose a simple \textit{soft} counting method based on similarity thresholding, obtained using a sigmoid function ($\sigma$) as a soft approximation of the step function. We define a scalar affine transform for the similarity values (domain of the sigmoid function), which is composed of two learnable scalar variables: slope $a$ and offset $b$. Thus, the weight $w_i$ for a local descriptor is obtained as a soft count,

\begin{equation}
    w_i = \sum_{j}\sigma(\mathit{a}d_{ij}+\mathit{b})
    \label{eq:weight}
\end{equation}
where $d_{ij}$ is computed as a dot product of features $\mathbf{x}_i$ and $\mathbf{x}_j$.
Following NetVLAD benchmarking in~\cite{berton2022deep}, we L2-normalize individual local features before computing similarities. 

\subsection{Feature Aggregation} 
Given the burstiness weight per local feature, we then use it for normalization within the standard NetVLAD aggregation~\cite{arandjelovic2016netvlad}:
\begin{equation}
\mathbf{V}_k = \sum_i^N\frac{\alpha_{ik}}{w_i^p}(\mathbf{x}_i - \mathbf{c}_k) \quad \forall k \in [0,C)
\label{eq:vlad}
\end{equation}
where $\alpha_{ik}$ denotes the original $1\times1$ convolution based learnable soft assignment value, initialized with feature-to-centroid distances~\cite{arandjelovic2016netvlad,berton2022deep}. $p$ is used as another learnable scalar variable, inspired by the original visual burstiness work~\cite{jegou2009burstiness} where the authors found it to be useful in dampening the effect of high value of counts.
Overall, \cref{eq:vlad} uniquely adapts the original NetVLAD aggregation with our novel \textit{soft count} method to obtain distinctive image representations by discounting burstiness (see~\cref{sec:ablatBurst} for corresponding ablation studies).

\subsection{Pre-Pool Projection} Our proposed weighting is based on intra-feature comparisons (\cref{eq:weight}), that is, $N\times N$ dot products. Furthermore, NetVLAD aggregation (\cref{eq:vlad}) requires $N\times C$ residual vector computations. Both of these operate on $D$-dimensional local descriptors. In order to reduce the compute time for aggregation, we propose to project these local features into a low-dimensional space ($D'$) \textit{before aggregation} through a \textit{PCA-initialized} learnable linear projection $\mathbf{R}_{D\times D'}$ with bias $\mathbf{m}$. Specifically, we initialize $\mathbf{R}_{D\times D'}$ and $\mathbf{m}$ respectively using the rotation matrix and mean vector computed from PCA transformation. We used a subset of local features in the training images for this purpose. Our pre-pool projected features are given by

\begin{equation}
    \mathbf{x}_i' = (\mathbf{x}_i - \mathbf{m}) \cdot \mathbf{R}_{D\times D'}, 
    \label{eq:pca}
\end{equation}
where $\mathbf{x}_i'$ represents $D'$ dimensional local features used in aggregation instead of the original local features $\mathbf{x}_i$. We note that an obvious alternative for such projection is to use a randomly-initialized bottleneck layer~\cite{sandler2018mobilenetv2}. However, in the context of learnable VLAD aggregation based on finetuning of pretrained large models, a \textit{randomly-initialized} projection of the original local feature space adversely affects the subsequent feature-to-cluster assignment. Our experiments show that the proposed PCA-based initialization for pre-pool projection retains superior recall even in a substantially lower-dimensional space, as demonstrated in~\cref{sec:result_pca}.

\section{Experimental Setup}
In this section, we present training and evaluation details along with the summary of benchmark dataset characteristics.

\subsection{VPR Training}
\label{subsec:exp_training}
\subsubsection{Training Dataset, Loss \& Backbone}
We train \ourM{} using multiple training configurations, which broadly vary in terms of \texttt{Dataset-Loss-Backbone}. For comparisons against state-of-the-art methods, we use 
the GSV dataset~\cite{ali2022gsv,ali2023mixvpr} with multi-similarity loss~\cite{wang2019multi} and DINOv2 (ViT-B)~\cite{oquab2023dinov2} backbone. We only fine-tune its last four layers, as proposed in SALAD~\cite{izquierdo2023optimal}. We use additional training configurations to extensively study the benefit of \ourM{} over vanilla VLAD (i.e., NetVLAD~\cite{arandjelovic2016netvlad}), as detailed in~\cref{sec:ablatBurst}. Both our proposed \ourM{} aggregation (\cref{eq:vlad}) and pre-pool PCA-initialized projection (\cref{eq:pca}) are defined on top of the last layer of backbone. For initializing the \ourM{} cluster centers (vocabulary $C=64$) and the pre-pool PCA transform, we randomly sample $50K$ features from the pretrained backbone using a diverse set of images across the training dataset. Following the original NetVLAD~\cite{arandjelovic2016netvlad}, we also use PCA whitening (PCAW) \textit{after} the aggregation to reduce the final descriptor size. In contrast, our \textit{pre-pool} PCA projection reduces the local descriptor size \textit{before} aggregation, which reduces \textit{both} the aggregation time and the final descriptor size.

\subsubsection{Burstiness Parameters} Our burstiness weighting layer involves three learnable parameters: $a$ (slope/temperature) and $b$ (offset)  to define the domain of sigmoid function, and $p$ to control the effect of soft count as a weight. We initialize $p$ with 1.0 and initialize $a$ and $b$ through a hyperparameter search on the Pitts30k validation set (as the default GSV training~\cite{ali2023mixvpr,izquierdo2023optimal} uses this for validation). 

\subsection{Evaluation} 
\subsubsection{Baselines} To ensure a fair and rigorous comparison with state-of-the-art, we benchmark \ourM{} against several leading methods, including CosPlace~\cite{berton2022rethinking}, MixVPR \cite{ali2023mixvpr}, EigenPlaces~\cite{berton2023eigenplaces}, AnyLoc~\cite{keetha2023anyloc}, and SALAD~\cite{izquierdo2023optimal}. MixVPR is trained using the GSV dataset~\cite{ali2023mixvpr}, whereas CosPlace and EigenPlaces are trained on the SFXL dataset~\cite{berton2022rethinking}. These methods represent the most recent and high-performing baselines reported in the VPR literature. 
For a fair comparison against SALAD, we also include results for SALAD with PCA Whitening.
Unlike SALAD, we do not use the \texttt{[CLS]} global token, thus it makes our method generally applicable to non-ViT backbones (e.g., see results in~\cref{tab:ablate_all_config} using a CNN backbone). Furthermore, we do not use any dustbin clusters, as used in SALAD and originally proposed in GhostVLAD~\cite{zhong2019ghostvlad}. While both the \texttt{[CLS]} global token and `ghost'/dustbin clusters can potentially further improve performance for \ourM{}, we leave it to future studies. 

\subsubsection{Benchmark Datasets} We used a significantly diverse set of standard public benchmark datasets for evaluation. This includes Pitts250k~\cite{toriivprrepet,arandjelovic2016netvlad} and Tokyo247~\cite{torii201524}, relevant to camera localization under strong viewpoint variations, along with day-night appearance variations in Tokyo247. We include MSLS (val)~\cite{warburg2020mapillary}, St Lucia~\cite{milford2008mapping}, SPED~\cite{chen2017deep} and Nordland~\cite{sunderhauf2013we} datasets relevant to robot navigation. We further include San Francisco SMall (SFSM)~\cite{berton2022rethinking} -- this small version of SFXL provided by~\cite{berton2022rethinking} is challenging in terms of viewpoint variations but can be evaluated rapidly. Furthermore, unlike the previous state-of-the-art method CosPlace which was trained on SFXL data, SFSM is still useful in testing generalization for GSV-based training. These datasets exhibit strong appearance changes, including moderate lighting variations of morning vs noon (with shadows) in St Lucia; different camera types, viewpoints and times of day in MSLS; and extreme variations due to seasonal change (winter vs summer) in Nordland. We further include AmsterTime and Baidu Mall datasets in our benchmarking to test generalization of our method. Both these datasets are relatively out of domain for the typical outdoor training common in VPR literature. AmsterTime compares images from across decades of time difference: historic vs recent street-view images. On the other hand, Baidu Mall is a highly-aliased indoor mall environment. Furthermore, our evaluation of \ourM{} is large-scale; these datasets vary in terms of their reference database size with around 76K, 27K and 18K images in Tokyo247, Nordland and MSLS respectively. We used~\cite{berton2022deep} to reformat these datasets and provide additional details in the supplementary. 

\subsubsection{Evaluation Metrics \& Preprocessing} We use Recall@K (K $\in \{1,5\}$) as our evaluation metric, considering VPR as a coarse initial retrieval step to facilitate the application of more precise metric localization and mapping techniques~\cite{tsintotas2022revisiting,garg2021where,masone2021survey,zaffar2021vpr,berton2022deep}. Ground truth reference images for queries are defined within a 25m localization radius for all datasets~\cite{berton2022deep}. 
We resize images to $224\times224$ during training  and $322\times322$ during evaluation for \ourM{} and SALAD -- this choice exactly follows the use of DINOv2 backbone in SALAD~\cite{izquierdo2023optimal}. Further preprocessing details for other baselines and datasets can be found in the supplementary.

\section{Results and Discussion}
In this section, we present state-of-the-art benchmark comparisons, followed by ablation studies to show the benefits of proposed pre-pool PCA projection, burstiness weighting and alternative feature weighting mechanisms. Finally, we provide qualitative examples to illustrate how burstiness weighting of \ourM{} upweights relevant local features for aggregation.

\begin{table*}[!t]
    \centering
    \caption{Recall@K comparison of our \ourM{} against state-of-the-art VPR techniques on challenging benchmark datasets with the best recall \textbf{bolded} and second-best \underline{underlined}.}    
    \resizebox{\linewidth}{!}{
    \begin{tabular}{l c c c c c c c c c c c c c c c c c c c c c c c c c c c c c c}
    \toprule
    \multirow{2}{*}{Method (Dim)} &&\multicolumn{2}{c}{MSLS}&&\multicolumn{2}{c}{NordLand}&&\multicolumn{2}{c}{Pitts250k-t}&&\multicolumn{2}{c}{SPED}&&\multicolumn{2}{c}{SFSM}&&\multicolumn{2}{c}{Tokyo247}&&\multicolumn{2}{c}{StLucia}&&\multicolumn{2}{c}{AmsterTime}&&\multicolumn{2}{c}{Baidu}\\
    \cline{3-4}\cline{6-7}\cline{9-10} \cline{12-13} \cline{15-16} \cline{18-19}\cline{21-22}\cline{24-25}\cline{27-28}\cline{30-31}
    &  T(ms) & R@1 & R@5 && R@1 & R@5 && R@1 & R@5 && R@1 & R@5 && R@1 & R@5&& R@1 & R@5 && R@1 & R@5&& R@1 & R@5 && R@1 & R@5\\
    \midrule
        CosPlace~\cite{berton2022rethinking} (512) & - & 79.5 & 87.2 && 51.3 & 66.8 && 89.7 & 96.4 && 79.6 & 90.4 && 78.1 & 84.8 && 87.0 & 94.9 && 99.0 & 99.9 && 47.7 & 69.8 && 41.6 & 55.0\\
        MixVPR~\cite{ali2023mixvpr} (4K) & - & 88.2 & 93.1 && 55.0 & 70.4 && 94.3 & 98.2 && 84.7 & 92.3 && 76.5 & 83.7 && 86.0 & 91.7 && 99.6 & {100} && 40.2 & 59.1 && 64.4 & 80.3\\
        EigenPlaces~\cite{berton2023eigenplaces} (2K) & - & 87.4 & 93.0 && 64.4 & 79.6 && 92.8 & 97.4 && 82.2& 93.7 && 84.6& 87.9&& 87.9& 94.9&& 99.5 & 99.9 && 49.8 & 72.4 && 61.5 & 76.4 &&\\
        AnyLoc~\cite{keetha2023anyloc} (49K) & - & 59.2 & 76.6 &&  9.6 & 16.3 && 88.1 & 95.5 && 76.6 & 90.0 && 70.4 & 79.9 && 85.7 & 94.3 && 79.0 & 90.4 && 41.3 & 61.8 && \underline{75.2} & \underline{87.6}\\

        SALAD~\cite{izquierdo2023optimal} (8K) & 14.6 & \underline{91.9} & \underline{95.8} && 70.8 & 85.4 && 94.9 & \underline{98.7} && 90.1 & 95.9 && 86.0 & 88.9 && 95.2 & 97.8 && \underline{100} & {100} && \underline{59.8} & \underline{79.2} && 69.1 & 82.5\\

        SALAD + PCAW (8K) & 14.6 & 90.3 & 95.3 && 69.0 & 83.6 && 94.7 & 98.6 && 89.6 & 95.1 && 83.8 & 87.1 && 95.6 & 97.8 && 99.9 & 99.9 && 52.2 & 70.0 && 70.7 & 84.0\\
        
    \midrule

        \textbf{\ourM{} (Ours):} & && && && && && && && && && \\

        w. PCAW (8K) & 44.8 & \textbf{92.4} & \underline{95.8} && \textbf{78.0} & \textbf{90.4} && \textbf{95.6} & \textbf{98.7} && \textbf{92.8} & \textbf{96.2} && \textbf{88.3} & \textbf{91.0} && \underline{96.5} & \underline{98.1} && \textbf{100} & {100} && \textbf{61.4} & \textbf{81.9} && \textbf{77.5} & \textbf{87.9}\\
        
        w. PrePool + PCAW (4K) & 14.2 & \underline{91.9} & \textbf{95.9} && \underline{71.4} & \underline{86.3} && \underline{95.0} & 98.2 && \underline{90.9} & \underline{96.0} && \underline{87.3} & \underline{90.1} && \textbf{97.5} & \textbf{98.4 }&& 99.9 & {100} && 59.4 & 78.7 && 74.3 & 86.6
        \\
    \bottomrule
    \end{tabular}}

\label{tab:sota_new}
\end{table*}

\subsection{State-of-the-art Benchmark Comparison}
In \cref{tab:sota_new}, we benchmark \ourM{} against recent top-performing methods. It can be observed that \ourM{} sets a new state-of-the-art by surpassing all prior methods across the board for both R@1 and R@5.
In comparison to AnyLoc, which uses a frozen DINOv2 backbone with hard VLAD assignment, our method surpasses it by large margins on all the datasets.
SALAD~\cite{izquierdo2023optimal} has recently shown that finetuning DINOv2 with Sinkhorn based aggregation significantly improves recall over MixVPR which uses the same training dataset/loss (GSV / MultiSim(MS))) but a different backbone and aggregation. Contrasting our proposed \ourM{} with SALAD, with only difference being the aggregation, we \textit{consistently} outperform SALAD, even without the use of \texttt{[CLS]} token or any dustbin clusters. As argued in AnyLoc, methods trained on street-view imagery (i.e., every method in the table except AnyLoc) typically generalize poorly in indoor environments. On two out-of-distribution datasets: AmsterTime (matching historic greyscale images with modern time RGB) and Baidu Mall (indoor), \ourM{} achieves much better generalization than SALAD's aggregation, which highlights the benefits of both VLAD-based aggregation and burstiness-based weighting. Finally, we report results for \ourM{} + Pre-pool PCA based aggregation (last row in~\cref{tab:sota_new}). Here, the local features are projected from $768$ to $192$ dimensions and the output VLAD vector of $12288$ ($=192\times 64$ clusters) is projected after PCAW to $4096$ dimensions. Overall, this reduces \ourM{}'s aggregation time from $44.8$ ms to $14.2$ ms (see~\cref{fig:ablation_1}), but retains a similar performance as \ourM{} even at half the final descriptor size while still outperforming SALAD.

\subsection{Ablation Studies}
In this section, we study the effect of pre-pool projection, burstiness weighting, different training configurations and alternative weighting methods to clearly emphasize the importance of our contributions.
\subsubsection{Pre-Pool Projection}
\label{sec:result_pca}
In \cref{fig:ablation_1}, we compare three different options for dimension reduction of local features \textit{before} aggregation: i) our proposed PCA-initialized linear transformation (\texttt{PCA init}), ii) randomly-initialized linear transformation (\texttt{Rand init Linear}), and iii) randomly-initialized Multi-Layer Perceptron (\texttt{Rand init MLP}, as Linear-ReLU-Linear). In all three cases, these projection layers are learnable within end-to-end training of \ourM{}. The $768$-dim result of \ourM{} does not use any pre-pool projection and serves as a reference. We also include results for SALAD by varying the output dimensions of its MLP-based pre-projection. We report Recall@1 on the MSLS (val set) against local feature dimensions (horizontal axis) and compute time (point size and as annotated in milliseconds), thus providing insights into the trade-off across recall, memory and compute time efficiency of aggregation. 
Here, compute time only accounts for feature projection and aggregation, and is measured using PyTorch Benchmark Utils (version 1.10.2) over 100 runs on an Nvidia RTX A5000. It can be clearly observed that \texttt{PCA-Init} maintains high recall even when the local feature dimensions are reduced $12\times$ from $768$ to $64$. \texttt{Rand Init Linear} is always inferior to \texttt{PCA init}, and the gap further increases for lower dimensions. On the other hand, \texttt{Rand Init MLP} suffers a major performance drop, which shows that including non-linearity or using additional layers does not help. Although SALAD initializes its pre-projection layers randomly~\cite{izquierdo2023optimal}, it behaves differently from the baseline \texttt{Rand init MLP} due to specific differences in cluster initialization and assignment procedures of VLAD and SALAD. For SALAD, recall drops at higher dimensions, and its aggregation time is roughly $50\%$ higher than \ourM{} at both the ends of the curve (see~\cref{tab:sota_new} for detailed results at a comparable operating point).
Overall, this study shows that high performance of \ourM{} can be maintained with significantly reduced requirements of compute time and memory by simply using PCA-based initialization.

 \begin{table}[t]
     \centering
    \begin{tabular}{c@{\hspace{0.3cm}}c}
        \begin{minipage}{0.49\textwidth}
            \centering
            \includegraphics[width=\textwidth]{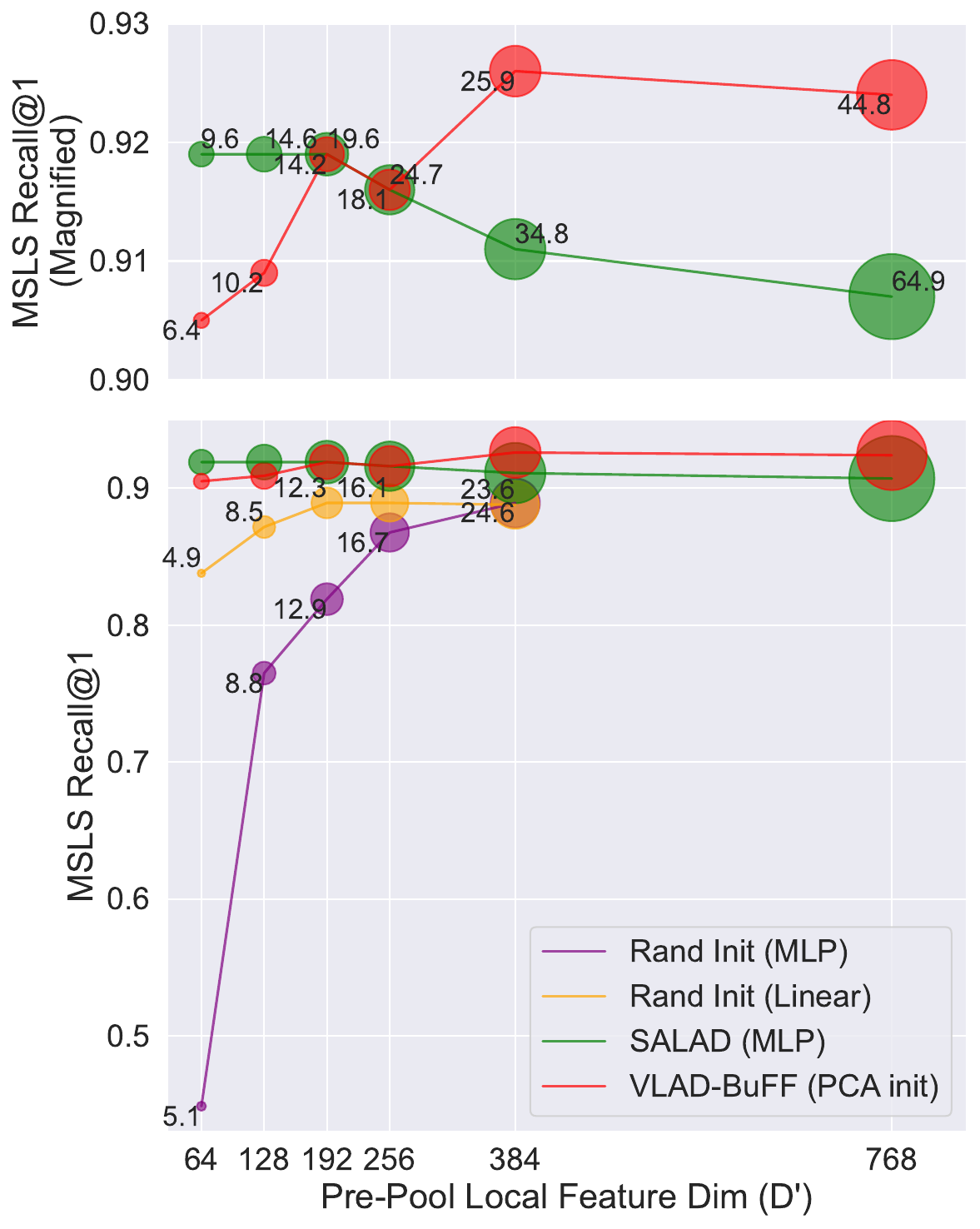}
            \captionof{figure}{Recall@1 for different \textit{Pre-pool Projection} techniques aimed at reducing local feature dimensions from $D=768$ to $D'=\{384, 256, 192, 128, 64\}$, with point size and annotations representing compute time for aggregation in milliseconds. Our proposed PCA-Init projection demonstrates superior recall-speed trade-off.}
            \label{fig:ablation_1}
        \end{minipage} &
        \begin{minipage}{0.49\textwidth}
            \centering
            \captionof{table}{MSLS Recall@1/5/10 across different training configurations, comparing our proposed burstiness-discounted weighting (\ourM{}) with a soft-assignment-only weighting (VLAD). Our method consistently improves recall across all training configurations.}   
            \begin{adjustbox}{width=\textwidth}
    \begin{tabular}{l c c c c}
    \hline
    \hline
    \multirow{2}{*}{Method (Dataset-Loss-Backbone)} && \multicolumn{3}{c}{MSLS}\\
    \cline{3-5}
    &  Dim & R@1 & R@5 & R@10\\
    \hline
    \textit{\textbf{a) P30K-Trip-R18}}:\\
    w. VLAD & 32K & 46.8 & 56.4 & 60.5 \\
    w. \ourM{} & 32K & \textbf{50.0} & \textbf{60.0} & \textbf{62.9} \\
    \hline
    \textit{\textbf{b) P30K-Trip-DINOv2}}:\\
    w. VLAD & 49K & 71.3 & 81.7 & 84.5 \\
    w. \ourM{} & 49K & \textbf{72.8} & \textbf{82.3} & \textbf{85.7} \\    
    \hline
    \textit{\textbf{c) SFXL-Clas-VGG16}}:\\
    w. VLAD & 512 & 80.1 & 88.2 & \textbf{90.5}\\
    w. \ourM{} & 512 & \textbf{80.3} & \textbf{88.3} & 90.4 \\  
    \hline
    \textit{\textbf{d) SFXL-Clas-R101}}:\\
        w. VLAD & 2K & 85.9 & 91.1 & 93.6   \\
        w. \ourM{} & 2K & \textbf{86.0} & \textbf{92.2} & \textbf{93.7} \\
    \hline               

    \textit{\textbf{e) GSV-MS-DINOv2 (PCAW)}}:\\
    w. 128-PrePool VLAD & 4K & 90.7 & 95.4 & 96.5 \\
    w. 128-PrePool \ourM{} & 4K & \textbf{90.9} & \textbf{95.7} & \textbf{96.2} \\    
    \hline               
    w. 192-PrePool VLAD & 4K & 91.6 & 95.7 & 96.6 \\
    w. 192-PrePool \ourM{} & 4K & \textbf{91.9} & \textbf{95.9} & \textbf{96.9} \\    
    \hline               
    w. 256-PrePool VLAD & 4K & 90.4 & 95.3 & 95.9 \\
    w. 256-PrePool \ourM{} & 4K & \textbf{91.6} & \textbf{96.2} & \textbf{96.8} \\    
    \hline               
    w. 384-PrePool VLAD & 4K & 92.0 & 96.1 & 96.5 \\
    w. 384-PrePool \ourM{} & 4K & \textbf{92.6} & \textbf{96.2} & \textbf{96.6} \\    
    \hline               
    % \textit{\textbf{   GSV-MS-DINOv2 (PCAW)}}:\\
    w. VLAD & 8K & 91.8 & \textbf{96.2} & 96.2 \\
    w. \ourM{} & 8K & \textbf{92.4} & 95.8 & \textbf{96.9} \\  

    \hline

    \end{tabular}
\end{adjustbox}
\label{tab:ablate_all_config}

        \end{minipage}  
    \end{tabular}
 \end{table}

\subsubsection{Burstiness Weighting}
\label{sec:ablatBurst}
In \cref{tab:ablate_all_config}, we compare the contribution of our burstiness-discounted weighting on top of the soft assignment weights used originally in NetVLAD. More precisely, this study evaluates \cref{eq:vlad} with and without the denominator $w_i^p$. To highlight the generic applicability and recall gains of our proposed weighting, we expand this study to several additional training configurations (\texttt{Dataset-Loss-Backbone}). \textit{a)}~\texttt{P30K-Trip-R18}, provided by the recent deep visual geolocalization benchmark~\cite{berton2022deep}: this closely corresponds to the original NetVLAD method trained on Pitts30k dataset with triplet loss~\cite{arandjelovic2016netvlad} but uses ResNet18 backbone for a better recall-speed trade-off. \textit{b)}~\texttt{P30K-Trip-DINOv2}, which is the same as the previous but replaces the ResNet18 backbone with DINOv2 to observe performance gains attributed to this powerful encoder alone. \textit{c)}~\texttt{SFXL-Clas-VGG16}, based on recent state-of-the-art method CosPlace~\cite{berton2022rethinking} which uses a cosine loss~\cite{wang2018cosface}, casting VPR problem as that of classification by grouping together images with similar heading and positions, and training on a very large dataset (SFXL). We drop-in replace CosPlace’s GeM~\cite{radenovic2018fine} pooling with VLAD and \ourM{} but retain its final fully connected layer that projects (GeM / VLAD / \ourM{}) descriptor to a fixed-size vector (512 in this case). \textit{d)}~\texttt{SFXL-Clas-R101}, which replaces the VGG16 backbone with ResNet101 and outputs 2048-dimensional descriptor after the final fully-connected layer. 
\textit{e)}~\texttt{GSV-MS-DINOv2}, which is our default training method with or without our pre-pool projection + PCAW, as described in~\cref{subsec:exp_training}. 

\cref{tab:ablate_all_config} shows that burstiness-discounting weighting \textit{consistently} improves recall across all different training configurations. These results highlight that the superiority of \ourM{} is not tied to a specific backbone, dataset or loss function. While the gain margins diminish for the bigger and better trained models, it is clear that innovations in aggregation methodologies still play a complementary role leading to superior performance. This is particularly important for resource-constrained robots, for example, in topological navigation~\cite{garg2024robohop,shah2023gnm,suomela2023placenav}, where high-performing but efficient visual place recognition is highly desirable.

\begin{table}[t]
    \centering
    \caption{Recall@1 comparison between \ourM{} and alternatives to feature weighting.}
\begin{adjustbox}{width=0.75\textwidth}
    \begin{tabular}{lccccccc}
    \toprule
     Method & $w_i$ Type & Train Features & SoftAssign & Tokyo247 & MSLS  \\
    \midrule
    No Weight & None & Y & Y & \textit{57.78} & 46.74\\
    PixelMLP & PixelMLP & Y & Y & 57.14 & 48.09 \\
    No-Train Weight & Burstiness & N & Y & 57.46 & \textit{48.80}\\ 
    \textbf{\ourM{}} (ours) & Burstiness & Y & Y & \textbf{59.37} & \textbf{49.97} \\
    \bottomrule
    \end{tabular}
\end{adjustbox}
    \label{tab:ablateAltWeights}
\end{table}

\subsubsection{Alternative Design Choices}
\label{sec:alternative}
We considered different design choices for feature weighting as alternatives to our proposed burstiness-discounted weighting. a) \textit{No Weight} -- this resembles the vanilla NetVLAD which uses soft assignment weighting but not the burstiness weighting. b) \textit{PixelMLP} -- we use an MLP with one hidden layer of size 64 and scalar output, which is applied to each local feature independently; we then apply softmax to the scalar values across all local features to obtain a relative weight per feature; this weight is then used as a replacement for $w_i$ in \cref{eq:vlad}. This style of weighting closely resembles attention-based weighting of TransVPR~\cite{wang2022transvpr} (which uses a single projection layer followed by softmax). c) \textit{No-Train Weight} -- we use the slope and offset parameters of our proposed weighting \textit{but do not train backbone's local features} with respect to the burstiness weighting layer. This represents a method analogous to the classical burstiness works~\cite{jegou2009burstiness} where local feature similarity is post-processed and thresholded to obtain feature weights without the option of fine-tuning local features through backpropagation. For these studies, we used the ~\texttt{P30K-Trip-R18} training recipe described in the previous subsection and report R@1 on MSLS (val) and Tokyo247 datasets.
In \cref{tab:ablateAltWeights},
it can be observed that PixelMLP outperforms vanilla NetVLAD on MSLS, but it is significantly behind the proposed burstiness-discounted weighting of \ourM{}. This highlights that learning a saliency weight by only observing the local descriptor itself is not sufficient. A similar trend is observed for No-Train Weight, which shows that learning local features in response to burstiness weighting is necessary for achieving superior performance.

\begin{figure}[t]
    \centering
    \begin{tabular}{cc}
    \includegraphics[width=0.5\textwidth]{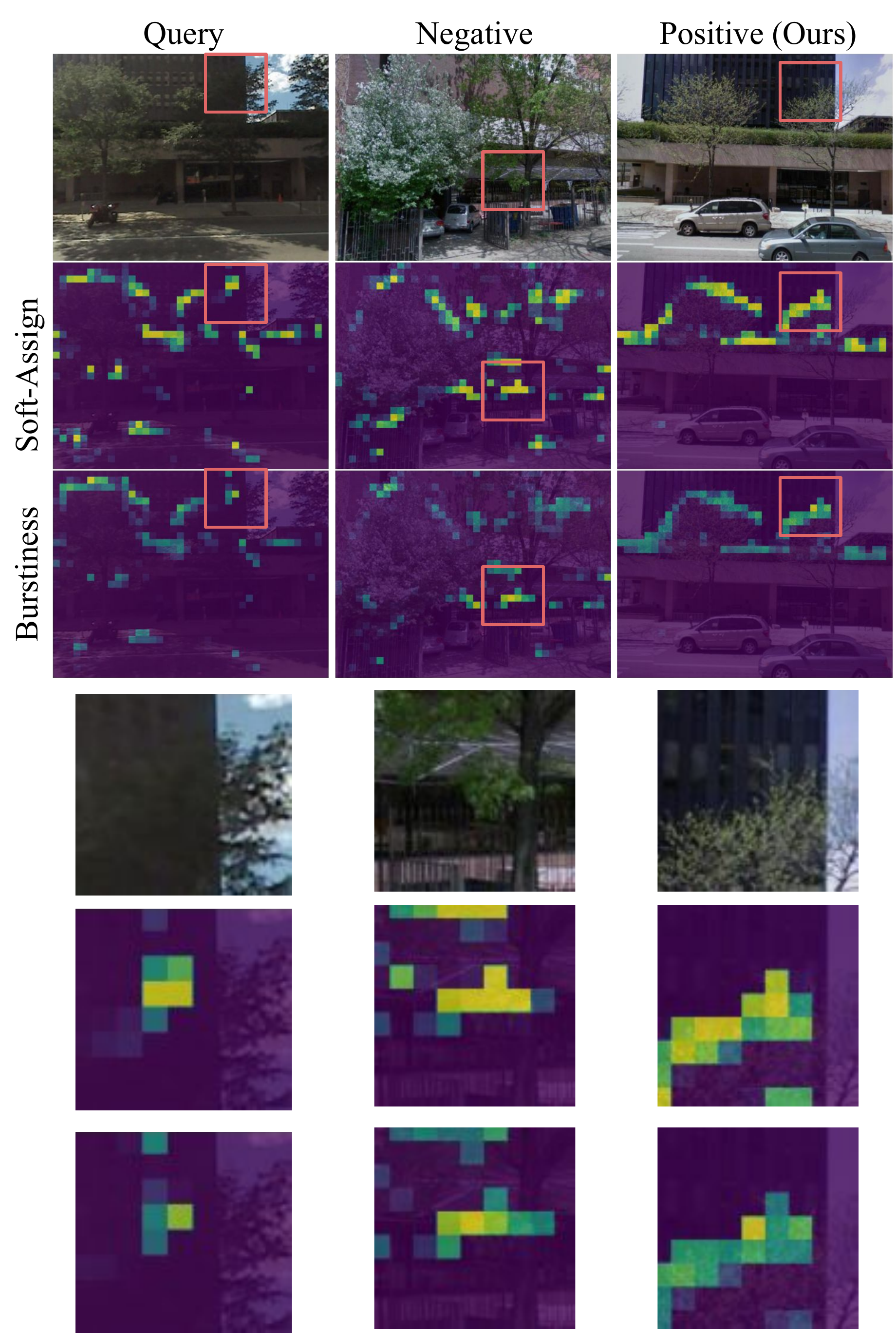} &
    \includegraphics[width=0.5\textwidth] {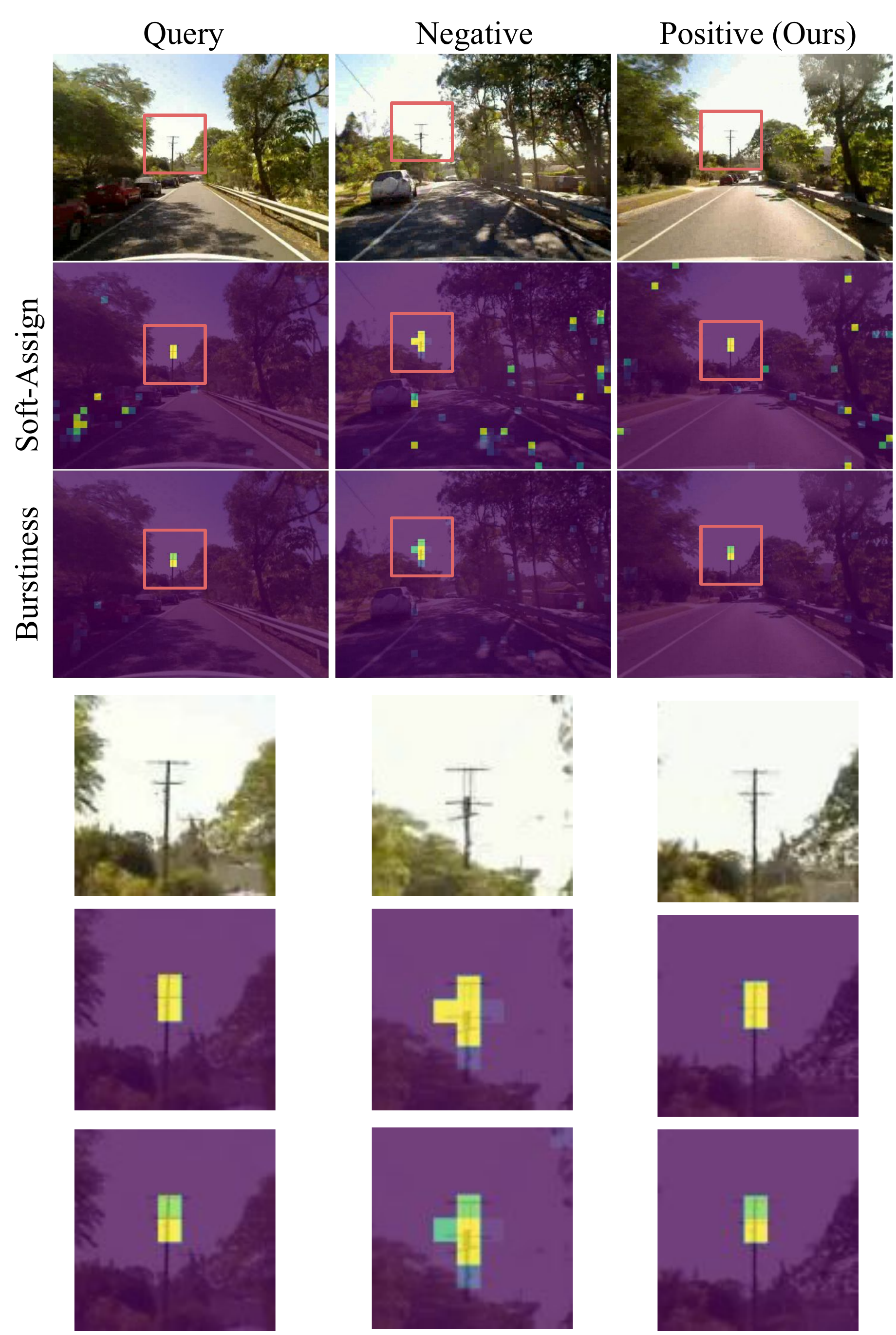} \\
    Case 1: Pitts30k & Case 2: St Lucia \\
    \end{tabular}

    \caption{Qualitative Analysis: Columns represent a query image, an incorrect match (negative) selected by vanilla NetVLAD and a correct match (positive) selected by \ourM{}. Mask colors represent weight values with lowest starting in blue and increasing through green to yellow. The image patches (bottom) are scaled versions of the local features marked with red boxes (top). Our proposed burstiness weighting selects specific features while avoiding repetitive patterns, as opposed to vanilla soft-assignment weighting.}
    \label{fig:qual}
\end{figure}

\subsection{Qualitative Analysis}
\cref{fig:qual} presents qualitative comparison between `with' and `without' the burstiness-discounted weighting applied on top of feature-to-centroid soft assignment, using two different datasets (additional examples can be found in the supplementary). For convenience, we refer to the `without burstiness' version as vanilla NetVLAD from here. The columns represent a triplet as a query-negative-positive, where the negative refers to the incorrect match selected by vanilla NetVLAD and positive refers to the correct match selected by our method. The top row represents raw RGB images and subsequent rows show overlaid mask of local feature weights for a particular cluster; these weights correspond to soft assignment weights ($\alpha_{ik}$) of vanilla NetVLAD model in the second row and burstiness weights ($\alpha_{ik}/w_i^p$) of our method in the third row. We also mark certain high-weighted features with red boxes and show their scaled versions in the bottom part of the figure. The mask colors represent the weight values with lowest starting in blue and increasing through green to yellow. All masks are auto-scaled per image as only the relative weighting matters due to cluster-wise L2-normalization of the aggregated residual. The procedure to select the clusters for this analysis is described in the supplementary.

In Case 1 (Pitts30k), we analyze a cluster which mostly captures features around the edges of trees and parts of building. It can be observed that soft assignment based weighting (2nd row) results in several local features contributing with a similar weight to the residual aggregation. On the other hand, our proposed burstiness weighting (3rd row, red boxes) up-weights specific features, e.g., the features in the red box are similarly weighted high in both the query and the positive, whereas the dominant local feature in the negative (red box) is found on the parking shed. Furthermore, the features with lower weights (in green) have a more consistent spatial distribution between query-positive than query-negative. It can also be noticed that some of the features from the tree's shadow on the road tend to have high weights when using vanilla weighting scheme (2nd row), whereas our method (3rd row) reduces the weight of such perceptually-aliasing features relative to the more relevant features.

In Case 2 (St Lucia), similar to Case 1, soft assignment based weighting (2nd row) leads to high weighting for several features found on vehicles, pole and shadows. This causes the query to match more with the negative than the positive for vanilla NetVLAD. Our proposed weighting (3rd row) results in a high weighting of a particular pole (within red boxes) consistently across the triplet, but the exact spatial distribution and weighting still matches more closely between the query and the positive.

\section{Limitations}
One of the core limitations of our method is its mostly outdoor-only applicability. Recently, AnyLoc~\cite{keetha2023anyloc} emphasized the lack of generalization for outdoor-trained models on indoor and unstructured environments. Our benchmark included only two `out-of-distribution' datasets: AmsterTime and Baidu Mall, both of which clearly show large room for absolute recall improvement. It is also evident that state-of-the-art results for R@1/5 are largely dependent on huge (compute-intensive) backbones. This often undermines innovations across other aspects of the VPR problem, which can potentially reduce reliance on large backbones. While our extensive benchmarking and ablation studies emphasized \ourM{}'s recall improvement across a variety of backbones, an extended in-depth analysis of corresponding trade-off between recall and compute time (memory) will be needed to better judge its practical utility for tasks like topological navigation with limited on-board compute resources~\cite{suomela2023placenav}.     

\section{Conclusion and Future Work}
In this paper, we presented \textbf{\ourM{}}, which extends vanilla NetVLAD~\cite{arandjelovic2016netvlad} with two key contributions: \textit{i)} a soft count based feature weighting mechanism that addresses the \textbf{Bu}rstiness problem for the first time within an end-to-end learning pipeline, and \textit{ii)} pre-pool PCA projection of local features for \textbf{F}ast \textbf{F}eature aggregation. \ourM{} sets a new state of the art on several benchmark datasets. Our pre-pool projection reduces compute time for aggregation without compromising on recall. 

 In this work, we particularly enhanced the VLAD aggregation through the burstiness-discounted weighting, which is inspired by the pre-deep learning literature. In future, a more comprehensive study could also explore the role and need for burstiness weighting in other performant aggregation techniques such as GeM~\cite{radenovic2018fine}. The burstiness weighting of \ourM{} can potentially be directly extended to local feature matching (e.g., Patch-NetVLAD~\cite{hausler2021patchnetvlad}), multi-scale aggregation (e.g., SPE-VLAD~\cite{yu2019spatial} and MR-NetVLAD~\cite{khaliq2022multires}), and sequential aggregation (e.g., SeqVLAD~\cite{mereu2022learning}), or to similar techniques in the rapidly growing field of LiDAR point cloud based place recognition~\cite{uy2018pointnetvlad}. Finally, more elaborate inductive biases could possibly be introduced in representation learning methods by further revisiting the older literature, e.g., query-database or intra-database matching based feature weighting~\cite{jegou2009burstiness} and local coordinate systems~\cite{delhumeau2013revisiting}.

\clearpage  %

\chapter*{Supplementary Material}
In this supplementary document, we present  extended results, additional implementation and background details for the proposed method, details of different training configurations, dataset details, abbreviations, and additional qualitative analyses.

\section{Extended Results}
Here, we provide additional results for different preprocessing and training configurations of the baseline methods and our proposed method.

\paragraph{\textbf{Upgraded Baselines:}} In~\cref{tab:sota_new_supp}, we present an extended version of state-of-the-art comparisons (Tab. 1 of the main paper). 

In block
\textcolor{OliveGreen}{\textbf{A}}, we include results for the authors' provided ResNet-101 based CosPlace model with $2048$ dimensional descriptors, which is their best performing model as reported in their supplementary~\cite{berton2022rethinking}. We also retrained CosPlace using its default GeM as well as our proposed VLAD-BuFF aggregation. It can be observed that our proposed aggregation on top of CosPlace's pipeline improves recall for most of the datasets.

In blocks \textcolor{OliveGreen}{\textbf{B}} and \textcolor{OliveGreen}{\textbf{C}}, we include the respective authors' trained models for CosPlace, EigenPlaces, and MixVPR -- they all differ in terms of \texttt{dataset-loss-backbone} but provide a reference to compare against their other variants in~\cref{tab:sota_new_supp}.
 
We further include `backbone-enhanced' versions of they key baselines. CosPlace and EigenPlaces used ResNet-101 and ResNet-50 backbones with GeM pooling and SFXL training dataset using their respective proposed loss functions. In block \textcolor{OliveGreen}{\textbf{D}}, we include a newly trained model for EigenPlaces where we \textit{only} replace its ResNet backbone with DINOv2. It can be observed that this replacement improves EigenPlaces' performance (compared to block \textcolor{OliveGreen}{\textbf{B}}) but does not achieve state-of-the-art results. In block \textcolor{OliveGreen}{\textbf{F}}, we use \texttt{GSV-MS-DINOv2} training configuration (as used by SALAD and our method). To compare different aggregation methods, we include newly trained \texttt{GSV-MS-DINOv2-GeM} and \texttt{GSV-MS-DINOv2-MixVPR} models. Table~\ref{tab:sota_new_supp} shows that \texttt{GSV-MS-DINOv2-GeM} leads to an improved recall for MSLS, SPED, Tokyo247, and Baidu datasets, in comparison to \texttt{SFXL-Clas-R101/R50} based CosPlace and \texttt{SFXL-Clas-R50} based  EigenPlaces. For \texttt{GSV-MS-DINOv2-MixVPR}, performance marginally improved for MSLS, SPED, SFSM, Tokyo247, and AmsterTime but decreased for Baidu, Nordland, and Pitts250k. Nevertheless, all these aforementioned upgraded baselines still perform inferior to our proposed method \ourM{} across all the datasets\footnote{Note that models trained with \texttt{DINOv2} as a backbone use 224$\times$224 resolution for training and 322$\times$322 for evaluation. The only exception is MixVPR, which requires a fixed image resolution due to its specific pooling method, thus we use 224$\times$224 resolution for both training and evaluation of MixVPR.}. 

\begin{table*}[t]
    \centering
    \caption{Recall@K comparison of our \ourM{} against state-of-the-art VPR techniques on challenging benchmark datasets with the best recall \textbf{bolded} and second-best \underline{underlined}.}    
    \resizebox{\linewidth}{!}{
    \begin{tabular}{l c c c c c c c c c c c c c c c c c c c c c c c c c c c c}
    \toprule
    \multirow{2}{*}{Method} &&\multicolumn{2}{c}{MSLS}&&\multicolumn{2}{c}{NordLand}&&\multicolumn{2}{c}{P250k-t}&&\multicolumn{2}{c}{SPED}&&\multicolumn{2}{c}{SFSM}&&\multicolumn{2}{c}{Tokyo247}&&\multicolumn{2}{c}{StLucia}&&\multicolumn{2}{c}{AmsterTime}&&\multicolumn{2}{c}{Baidu}\\
    \cline{3-4}\cline{6-7}\cline{9-10} \cline{12-13} \cline{15-16} \cline{18-19}\cline{21-22}\cline{24-25}\cline{27-28}
    &  Dim & R@1 & R@5 && R@1 & R@5 && R@1 & R@5 && R@1 & R@5 && R@1 & R@5&& R@1 & R@5 && R@1 & R@5&& R@1 & R@5 && R@1 & R@5\\
    
    \midrule
        \textcolor{OliveGreen}{\textbf{A)}} \texttt{SFXL-Clas-R101:} & && && && && && && && && && \\   
        CosPlace (CP)$^\dagger$~\cite{berton2022rethinking} &  2048 &  90.4 & 94.2 &&  49.7& 66.0 &&   91.8& 95.6  && 76.4 & 85.5 && 81.9 & 86.6 && 86.7 & 95.6 &&  99.9 & 99.9  && 51.9 & 72.0 && 42.5 & 55.8\\    
        
        CP w. GeM & 2048& 86.2 &91.9 && 47.1&63.1 && 92.0&97.6 &&  76.3&88.6 && 80.2&85.9 &&  89.2&94.9 &&  99.2&99.9 && 51.7&72.1 && 41.7&54.2\\
        CP w. \ourM{} & 2048 & 86.0&92.2 && 57.7&73.6 && 92.7&98.0 && 78.6&90.0 && 80.3&86.3 && 89.2&96.2 && 99.5&99.9 && 52.9&72.6 && 40.8&54.8\\

        \midrule   
        \textcolor{OliveGreen}{\textbf{B)}} \texttt{SFXL-Clas-R50:} & && && && && && && && && && \\   
        CosPlace$^\dagger$~\cite{berton2022rethinking} &  512 & 79.5 & 87.2 && 51.3 & 66.8 && 89.7 & 96.4 && 79.6 & 90.4 && 78.1 & 84.8 && 87.0 & 94.9 && 99.0 & 99.9 && 47.7 & 69.8 && 41.6 & 55.0\\
        EigenPlaces$^\dagger$~\cite{berton2023eigenplaces} & 2048 & 87.4 & 93.0 && 64.4 & 79.6 && 92.8 & 97.4 && 82.2& 93.7 && 84.6& 87.9&& 87.9& 94.9&& 99.5 & 99.9 && 49.8 & 72.4 && 61.5 & 76.4\\
        \midrule
        \textcolor{OliveGreen}{\textbf{C)}} \texttt{GSV-MS-R50:} \\
        MixVPR$^\dagger$~\cite{ali2023mixvpr} & 4096 & 88.2 & 93.1 && 55.0 & 70.4 && 94.3 & 98.2 && 84.7 & 92.3 && 76.5 & 83.7 && 86.0 & 91.7 && 99.6 & 100 &&  40.2 & 59.1 && 64.4 & 80.3\\
        \midrule
        \textcolor{OliveGreen}{\textbf{D)}} \texttt{SFXL-Clas-DINOv2:} & && && && && && && && && && \\   
        EigenPlaces & 512 & 91.1 & 95.4 && 66.8 & 82.4 && \textbf{95.8} & \textbf{99.1} &&  86.2 & 95.1 && 86.5 & 89.9 && 92.7 & 97.1  &&  99.7 & 99.9 && 51.6 & 72.6 && 62.7 & 77.7\\
        \midrule
        \textcolor{OliveGreen}{\textbf{E)}} \texttt{DINOv2-Pretrained:} & && && && && && && && && && \\AnyLoc~\cite{keetha2023anyloc} & 49K & 59.2 & 76.6 &&  9.6 & 16.3 && 88.1 & 95.5 && 76.6 & 90.0 && 70.4 & 79.9 && 85.7 & 94.3 && 79.0 & 90.4 && 41.3 & 61.8 && 75.2 & \underline{87.6}\\
        \midrule
        \textcolor{OliveGreen}{\textbf{F)}} \texttt{GSV-MS-DINOv2:} & && && && && && && && && && \\    
        GeM & 2048 & 88.9 & 94.7 && 54.2 & 72.0&& 93.4 & 97.9&& 85.2 & 92.6 && 81.0 & 85.4 && 90.5 & 95.2 && 99.8 & 100 &&45.7 & 67.7 &&63.7 & 79.5\\     
        MixVPR & 4096 & 88.8 & 94.6 && 52.7 & 70.7 && 92.6 & 97.6 &&88.3 & 92.6&& 78.3 & 85.5&& 89.8 & 94.9 && 99.7 & 100 && 46.6 & 69.6 && 51.7 & 70.7\\
         SALAD$^\dagger$ & 8448 & 92.2 & \textbf{96.2} && 76.1 & \underline{89.2} && 95.1 & 98.5 && \underline{92.1} & \textbf{96.2} && 85.4 & 88.5 && 95.2 & 97.1 && 99.9 & 100 && 58.7 & 79.2 && 68.5 & 81.7\\
         
         SALAD$^\dagger$ + PCAW & 8192 & 90.3 & 95.3 && 74.5 & 87.5 && 94.8 & 98.3 && 91.3 & 95.6 && 83.9 & 87.4 && 93.3 & 97.5 && 99.9 & 100 && 52.2 & 71.3 && 67.2 & 80.9 \\
        
        SALAD & 8448 & 91.9 & 95.8 && 70.8 & 85.4 && 94.9 & \underline{98.7} && 90.1 & 95.9 && 86.0 & 88.9 && 95.2 & 97.8 && \textbf{100} & 100 && 59.8 & 79.2 && 69.1 & 82.5\\

        SALAD + PCAW & 8192 & 90.3 & 95.3 && 69.0 & 83.6 && 94.7 & 98.6 && 89.6 & 95.1 && 83.8 & 87.1 && 95.6 & 97.8 && 99.9 & \underline{99.9} && 52.2 & 70.0 && 70.7 & 84.0\\

        NetVLAD + PCAW & 8192 & 91.8 & \textbf{96.2} && \underline{76.2} & \underline{89.2} && \underline{95.7} & 98.6 && 91.3 & \textbf{96.2} && \underline{88.2} & \underline{90.4} && \underline{97.1} & \underline{98.1} && \underline{99.9} & 100 && \textbf{61.8} & \textbf{82.1} && \underline{76.4} & 86.7\\

        \textbf{\ourM{} (Ours):} & && && && && && && && && && \\
        w. PCAW & 8192 & \textbf{92.4} & 95.8 && \textbf{78.0} & \textbf{90.4} && 95.6 & \underline{98.7} && \textbf{92.8} & \textbf{96.2} && \textbf{88.3} & \textbf{91.0} && 96.5 & \underline{98.1} && \textbf{100} & 100 && \underline{61.4} & \underline{81.9} && \textbf{77.5} & \textbf{87.9}\\
        
        w. PrePool + PCAW & 4096 & \underline{91.9} & \underline{95.9} && 71.4 & 86.3 && 95.0 & 98.2 && 90.9 & \underline{96.0} && 87.3 & 90.1 && \textbf{97.5} & \textbf{98.4} && \underline{99.9} & 100 && 59.4 & 78.7 && 74.3 & 86.6
        \\

    \bottomrule
\multicolumn{5}{l}{$^\dagger$Using the original authors' provided model.}
    \end{tabular}}
\label{tab:sota_new_supp}

\end{table*}

\paragraph{\textbf{SALAD Variants}:}In our main paper, we reported results for SALAD~\cite{izquierdo2023optimal} model that we trained to ensure a fully fair comparison with our proposed method. 
In Table~\ref{tab:sota_new_supp}, we include results using the checkpoint provided by the authors ($^\dagger$). It can be observed that \ourM{} still remains the state-of-the-art method across the majority of the datasets.
It can be further observed that using WPCA on SALAD consistently reduces recall, which significantly limits its potential for efficient retrieval based on low dimensional global descriptors~\cite{berton2022deep,zaffar2021vpr,berton2022rethinking}. 

\paragraph{\textbf{Upgraded NetVLAD}:} We include an upgraded version of NetVLAD~\cite{arandjelovic2016netvlad} (based on \texttt{GSV-MS-DINOv2} training configuration) as another strong state-of-the-art baseline, which only differs from \ourM{} in terms of aggregation and is referred to as NetVLAD + PCAW in~\cref{tab:sota_new_supp}. It can be observed that VLAD aggregation, common in both NetVLAD and \ourM{}, is generally superior to all other baselines. \ourM{} improves this with the proposed burstiness weighting, achieving higher recall on the majority of datasets. Furthermore, our pre-pool projection based on PCA initialization retains superior performance with significantly reduced descriptor dimensions and aggregation compute time.

\section{Implementation Details}
Here, we provide background details for vanilla NetVLAD's soft assignment, followed by the description of parameter settings for our training configurations.

\subsection{Vanilla NetVLAD Soft Assignment}
Our proposed method~\ourM{} uses a burstiness weighting ($w_i$) based on feature-to-feature similarities. This is applied on top of feature-to-cluster assignment ($\alpha_{ik}$) to obtain intra-cluster burstiness weighting. Here, we provide the background details for $\alpha_{ik}$, originally proposed in NetVLAD~\cite{arandjelovic2016netvlad}. The authors implemented it as a convolutional layer with $1\times1$ spatial support and $|C|$ weight filters/kernels, followed by a softmax across the spatial dimensions of the input local feature tensor. The convolutional filters are the learnable parameters of the NetVLAD aggregation layer, where each filter is initialized with the cluster center $C_k$. Furthermore, the authors used a multiplication factor for these weight filters during initialization to mimic the sparse assignment of conventional VLAD, as described in the supplementary material of NetVLAD~\cite{arandjelovic2016netvlad} and detailed in their official implementation\footnote{https://github.com/Relja/netvlad/blob/master/README\_more.md} which also notes the preference of `vlad' over `vladv2' for L2-normalized features.

\subsection{Training and Evaluation Configurations}

\subsubsection{GSV-DINOv2}
\label{sec:gsv_dinov2}
For training and testing of GSV-DINOv2 pipeline, we used the official code\footnote{https://github.com/serizba/salad} of the recent state-of-the-art method SALAD~\cite{izquierdo2023optimal}. We substitute the SALAD aggregation layer with our \ourM{} aggregation (or other baseline aggregation methods such as MixVPR and GeM). We exactly follow SALAD's training parameters setting, including the Multi-Similarity loss, AdamW optimizer, learning rate $6\mathrm{e}{-5}$, and batch size $60$ (each comprising 4 images per place). Each training iteration is completed in approx. 45 minutes on an NVIDIA A100, spanning 4 epochs, which saves the best 3 models according to Recall@1 on Pitts30k-val set. The number of clusters is set to 64 for both \ourM{} and SALAD. While we employed image sizes of 224$\times$224 and 322$\times$322 for training and evaluation across most methods utilizing GSV-DINOv2 pipeline, MixVPR uniquely required 224$\times$224 dimensions for both due to its strict architecture needs. 

For our benchmarking, we used the evaluation codebases provided by~\cite{izquierdo2023optimal,berton2022deep,berton2023eigenplaces}\footnote{https://github.com/gmberton/VPR-methods-evaluation} to access the official model checkpoints. For the datasets, we used dataset downloader provided by~\cite{berton2022deep}\footnote{ https://github.com/gmberton/VPR-datasets-downloader} for AmsterTime; SPED and Nordland were sourced from VPR-Bench~\cite{zaffar2021vpr}; Tokyo247 and StLucia from DVGL~\cite{berton2022deep}; and Baidu from AnyLoc~\cite{keetha2023anyloc}.

\subsubsection{Deep Visual Geolocalization (DVGL) Benchmark}
We used the official code\footnote{https://github.com/gmberton/deep-visual-geo-localization-benchmark} for the recent deep visual geolocalization benchmark~\cite{berton2022deep} to train and test NetVLAD and \ourM{}. This corresponds to the ~\texttt{P30K-Trip-R18} (and ~\texttt{P30K-Trip-DINOv2}) configuration. Here, we provide details of the key train/test parameter settings which strictly follow the official code. For training, batch size is 4 where each sample is composed of 12 images: one query (anchor), one positive and ten negatives mined through a `partial' strategy~\cite{berton2022deep}. Triplet margin loss~\cite{balntas2016learning} is used with Euclidean distances and 0.1 margin, and optimized using Adam with a learning rate of 0.00001. The backbone network is ResNet18 which is clipped beyond conv4 and its pretrained ImageNet weights frozen, keeping only conv4 as trainable. We neither use PCA nor any supervised linear projection (fully connected layer) \textit{after} feature aggregation. All local features are L2-normalized before aggregation. No data augmentation is used for training. Images are resized to 640$\times$480 for both training and evaluation. An early stopping criterion is defined to track improvement in Recall@5 over 3 epochs, with maximum training epochs set to 15. The number of clusters for NetVLAD and \ourM{} are set to 64. For ~\texttt{P30K-Trip-DINOv2}, everything remains the same except that the backbone is used in the same way as that for GSV-DINOv2 training described earlier.

\subsection{Burstiness Exponent $p$}
For the burstiness weighting layer in \ourM{}, we used $p$ as an exponent for the soft count. This was initialized as 1.0. We observed that its final trained value converged between 0.7 and 0.8 for different training configuration models. This is in line with the findings of~\cite{jegou2009burstiness} where their best performing value was 0.5 (as opposed to 1.0), which our method converges towards.

\section{Datasets Details}

\textbf{Pittsburgh:}
The Pittsburgh dataset~\cite{toriivprrepet} consists 250k images collected using Google Street View. The dataset is divided into a number of places, with 24 images captured per place from different viewing directions (two pitch and twelve yaw directions). The reference and query images are captured at different times of the day and several years apart. In this work, we use the full dataset (Pitts250k) for testing and use its 30K image subset (Pitts30K)~\cite{arandjelovic2016netvlad} for training and validation. The Pitts250k-test set contains 8280 and 83952 images in query and reference traverses.

\noindent\textbf{Tokyo 24/7:}
The Tokyo 24/7 dataset~\cite{torii201524} contains a large database (75984 images) collected from Google Street View and a small set of query images collected using smartphones. The query traverse (315 images) are captured at 125 different locations with different viewing directions at different times of the day (including night time).

\noindent\textbf{Mapillary Street Level Sequences (MSLS):}
The MSLS dataset~\cite{warburg2020mapillary} consists of approximately 1.6 million images collected from 30 different cities across the world over seven years. The dataset is split into training, validation and test sets, with ground truth released for the training and validation sets. In this work, we evaluate our method on the validation set (740 and 18871 images for query and reference traverses), as per recent works~\cite{hausler2021patchnetvlad, berton2022deep}.

\noindent\textbf{St Lucia:}
The St Lucia dataset~\cite{milford2008mapping} was collected by a car driving around a fixed route of a suburb (St Lucia) in Brisbane, at different times of day. We utilize the subset of this dataset used in the deep visual geolocalization benchmark~\cite{berton2022deep}, which uses the temporally first and last drives along this route. The query and reference traverses consist of 1464 and 1549 images respectively.

\noindent\textbf{Nordland:}
The Nordland dataset~\cite{sunderhauf2013we} is recorded by a train traveling through Norway during four different seasons for 728km. We used the subset of the dataset as used in VPR-Bench~\cite{zaffar2021vpr}, where the winter route is used as the database (27K images) and the summer route is the query set (2.7K images).

\noindent\textbf{Baidu:}
The Baidu dataset~\cite{keetha2023anyloc} captured in a mall, includes images with varied camera poses and provides both the ground truth location and 3D pose for each image. It is suitable for 6-Degrees of Freedom (DoF) Localization and VPR testing, comprising 2292 query images and 689 reference images. The dataset is particularly challenging due to perceptual aliasing, dynamic objects (e.g., people), and semantically rich elements like billboards and signs, offering a comprehensive environment for VPR evaluation.

\noindent\textbf{AmsterTime:}
The AmsterTime dataset~\cite{yildiz2022amstertime} captures Amsterdam's urban/historic scenery at different times across decades, offering a unique challenge of recognizing the same location from different viewpoints and colorscale (RGB vs grey). The query and reference traverses consist of 1231 and 1231 images ~\cite{berton2023eigenplaces}. 

\noindent\textbf{San Francisco SMall:}
The San Francisco Small (SF-SMall) dataset~\cite{berton2022rethinking}, a subset of a larger collection SF-XL, focuses on the San Francisco area, providing a dense urban environment for VPR evaluation. The query and reference traverse consists of 1K and 27191 images.

\noindent\textbf{SPED:}
The SPED dataset~\cite{chen2018learning} is renowned for its diverse environmental conditions, showcasing substantial seasonal and illumination changes, with winter scenes as queries (607 images) and summer scenes as references (607 images). This variability, including moderate viewpoint shifts, makes SPED a challenging dataset for VPR.

\section{Abbreviations}
In ~\cref{tab:abbrevs}, we provide a list of abbreviations used in the main paper. 

\begin{table}
    \centering
    \caption{List of abbreviations used in this paper.}
\begin{adjustbox}{width=0.45\textwidth}
    \begin{tabular}{cc}
    \toprule
     Abbreviation & Description \\
    \midrule
    GAP & Global Average Pooling \\
    GeM & Generalized Mean Pooling \\
    GPO & Generalized Pooling Operator \\
    MAC & Maximum Activations of Convolution \\ 
    PCA & Principal Component Analysis \\
    ViT & Vision Transformer \\
    VLAD & Vector of Locally Aggregated Descriptors \\
    VPR & Visual Place Recognition \\
    \bottomrule
    \end{tabular}
    \label{tab:abbrevs}
\end{adjustbox}
\end{table}

\section{Qualitative Analysis}
Here, we provide additional qualitative examples that compare the weighting patterns between vanilla NetVLAD's soft assignment and \ourM{}'s burstiness, see \cref{fig:qual1}-~\cref{fig:qual4} with detailed captions. In \cref{fig:qual3} and~\cref{fig:qual4}, we show soft assignment weighting for \textit{both} NetVLAD and \ourM{} to highlight the impact of burstiness weighting on learning backend local features, resulting in noticeable variations in their cluster assignment. In ~\cref{fig:qual4}, we present a failure case of our method. We used the P30K-R18 trained models for these qualitative analyses; in the subsequent section, we describe our procedure to select the clusters for the visualization purpose.

\subsection{Procedure}
We select the triplet (query, positive and negative) using ground truth by obtaining query indices where \ourM{} performed better than vanilla NetVLAD based weighting. For each image, we compute its $C\times D$ VLAD representation which is L2-normalized (first cluster-wise and then over the flattened vector~\cite{arandjelovic2016netvlad}). For each normalized $D$-dimensional aggregated residual vector per cluster $C_k$, we compute a triplet margin $m_k$ as the difference between query-negative and query-positive Euclidean distances:
\begin{equation}
    m_k = ||\textbf{q}_k - \textbf{n}_k|| - ||\textbf{q}_k - \textbf{p}_k||
\end{equation}

For a cluster, a larger margin indicates its ability to better separate a negative from the query-positive pair. We use these margins to obtain the ranking of clusters for both the soft-assignment NetVLAD baseline and \ourM{} models, referred to as $r_k^{NV}$ and $r_k^{VB}$ respectively. We then compute cluster rank difference between the two models to obtain the cluster index which maximally changed the triplet margin, as below:
\begin{equation}
    \bar{k} = \operatorname{argmax}(r_k^{NV}-r_k^{VB})
\end{equation}

Intuitively, $\bar{k}$ will be selected as the cluster index which led to low margin in NV and high margin in VB, thus being crucial in match selection for the query.

\newcommand{\scaleone}{0.8\textwidth}
\begin{figure*}
    \centering
    \begin{tabular}{
    p{0.15cm}
    >{\centering\arraybackslash}p{3cm}
    >{\centering\arraybackslash}p{3cm}
    >{\centering\arraybackslash}p{3cm}
    }
    & Query & Negative & Positive (Ours) 
    \\
    
     & 
     \multicolumn{3}{c}{\includegraphics[width=\scaleone]{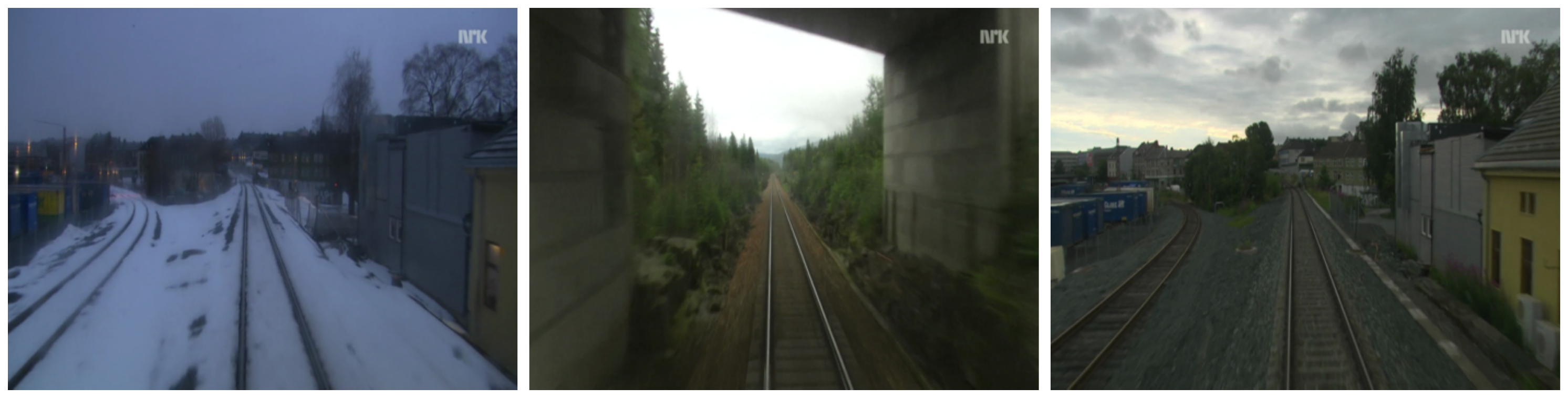}} 
     \\
     
     \rotatebox{90}{\hspace{0.5cm}Soft Assign} &
     \multicolumn{3}{c}{\includegraphics[width=\scaleone]{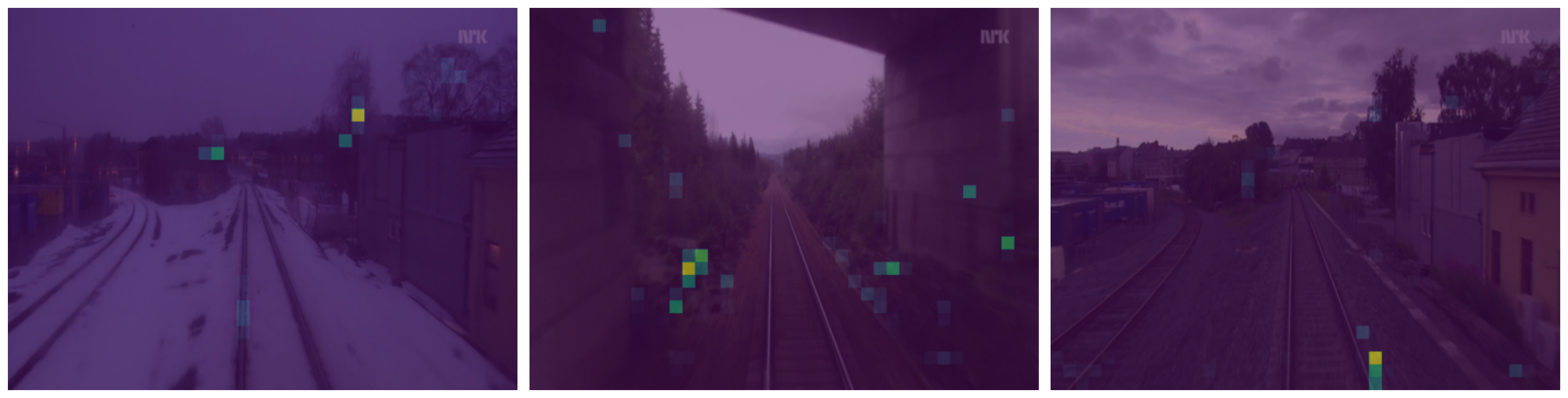}} 
     \\
     
     \rotatebox{90}{\hspace{0.5cm}Burstiness} &

     \multicolumn{3}{c}{\includegraphics[width=\scaleone]{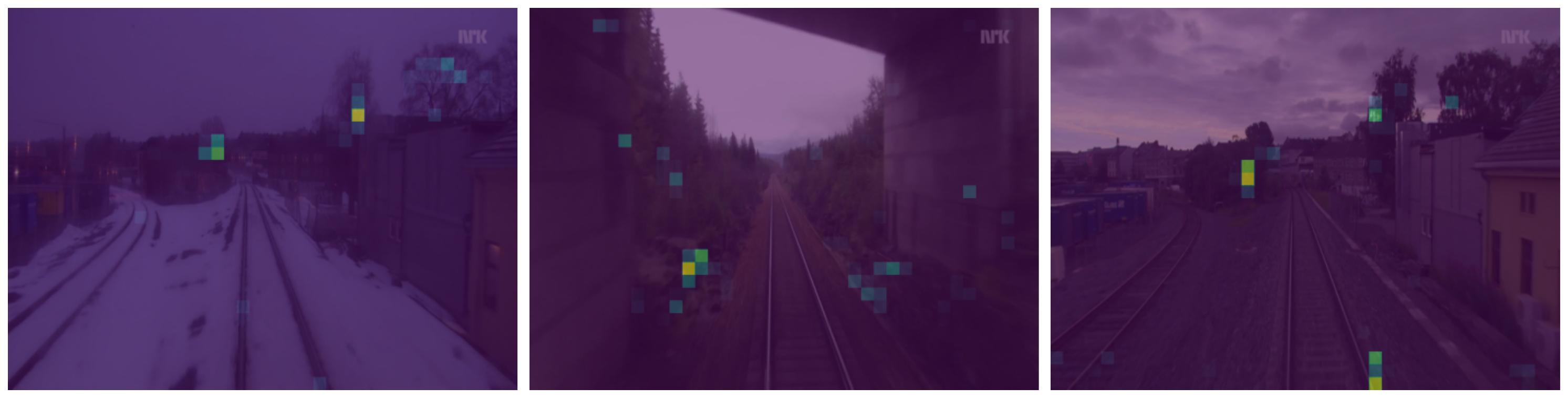}} 
     \\

    \end{tabular}
    \caption{In this example from the Nordland dataset, both the vanilla soft assignment (2nd row) and our proposed method (3rd row) select image regions lying just beneath the tree canopies (first column). In the positive image (third column), both the methods select some features on the railway tracks (highlighted in yellow). However, our method downweights these features relative to those below the tree canopies (more yellow than tracks), thus improving the query-positive matching.}
    \label{fig:qual1}
\end{figure*}

\begin{figure*}
    \centering
    \begin{tabular}{
    p{0.15cm}
    >{\centering\arraybackslash}p{3cm}
    >{\centering\arraybackslash}p{3cm}
    >{\centering\arraybackslash}p{3cm}
    }
    & Query & Negative & Positive (Ours) 
    \\
    &
     \multicolumn{3}{c}{\includegraphics[width=\scaleone]{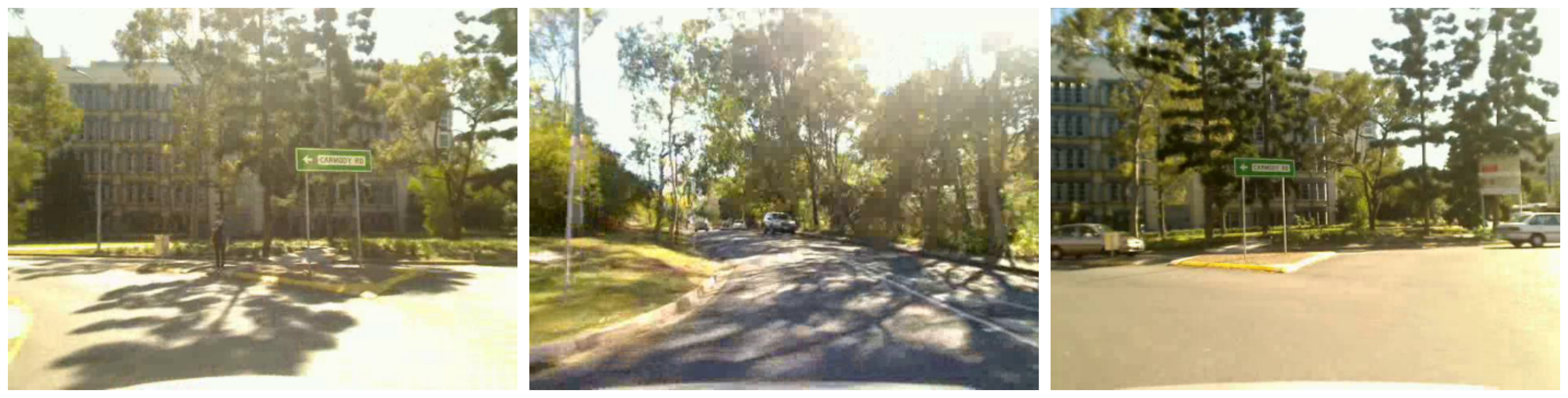}} 
     \\
     
     \rotatebox{90}{\hspace{0.5cm}Soft Assign} &

     \multicolumn{3}{c}{\includegraphics[width=\scaleone]{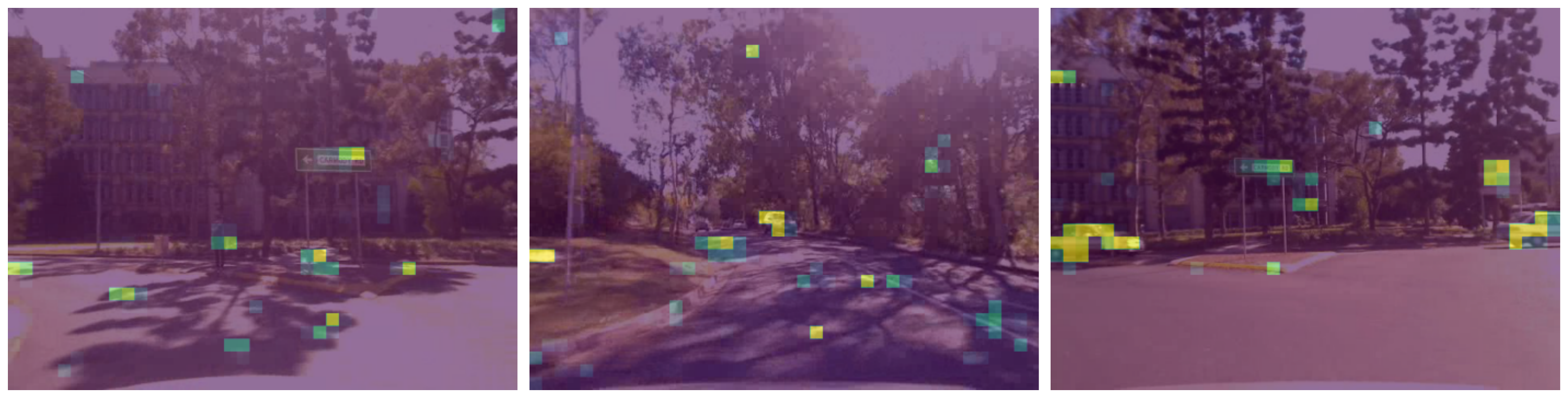}} 
     \\
     
     \rotatebox{90}{\hspace{0.5cm}Burstiness} &

     \multicolumn{3}{c}{\includegraphics[width=\scaleone]{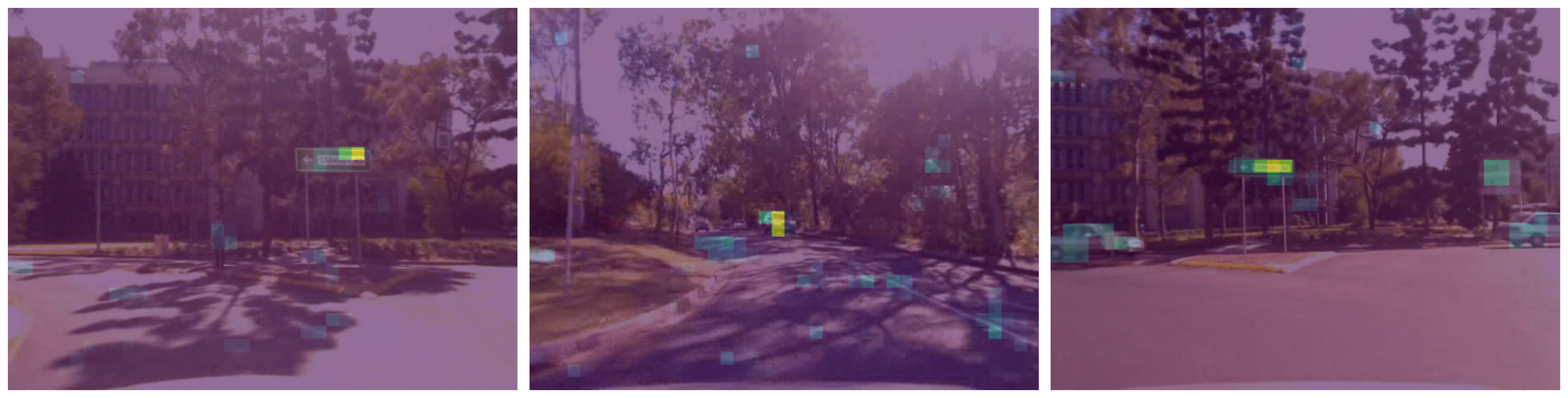}} 
     \\

    \end{tabular}
    \caption{In this example from the St Lucia dataset, vanilla soft assignment results in several highly-weighted features, including those on trees' shadows (query and negative) and vehicles (positive). However, our proposed weighting (3rd row) downweights all the features found on shadows and vehicles, and instead selects a signboard with a consistent weighting pattern between the query-positive pair.}
    \label{fig:qual2}
\end{figure*}

\begin{figure*}
    \centering
    \begin{tabular}{
    p{0.15cm}
    >{\centering\arraybackslash}p{3cm}
    >{\centering\arraybackslash}p{3cm}
    >{\centering\arraybackslash}p{3cm}
    }
    & Query & Negative & Positive (Ours) 
    \\
    &
     \multicolumn{3}{c}{\includegraphics[width=\scaleone]{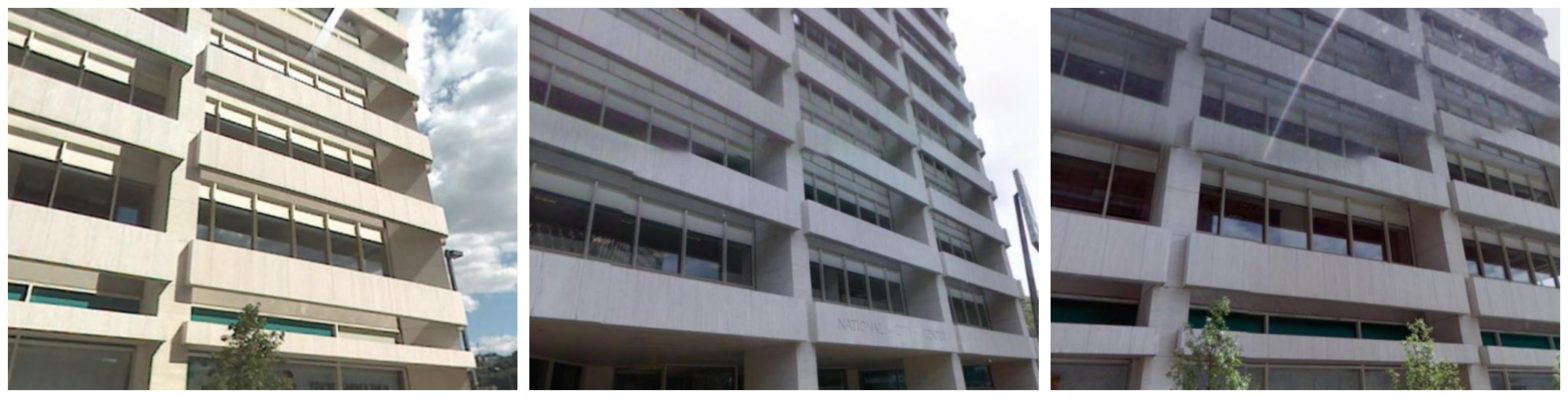}} 
     \\
     
     \rotatebox{90}{\hspace{0.5cm}SA (vanilla)} &

     \multicolumn{3}{c}{\includegraphics[width=\scaleone]{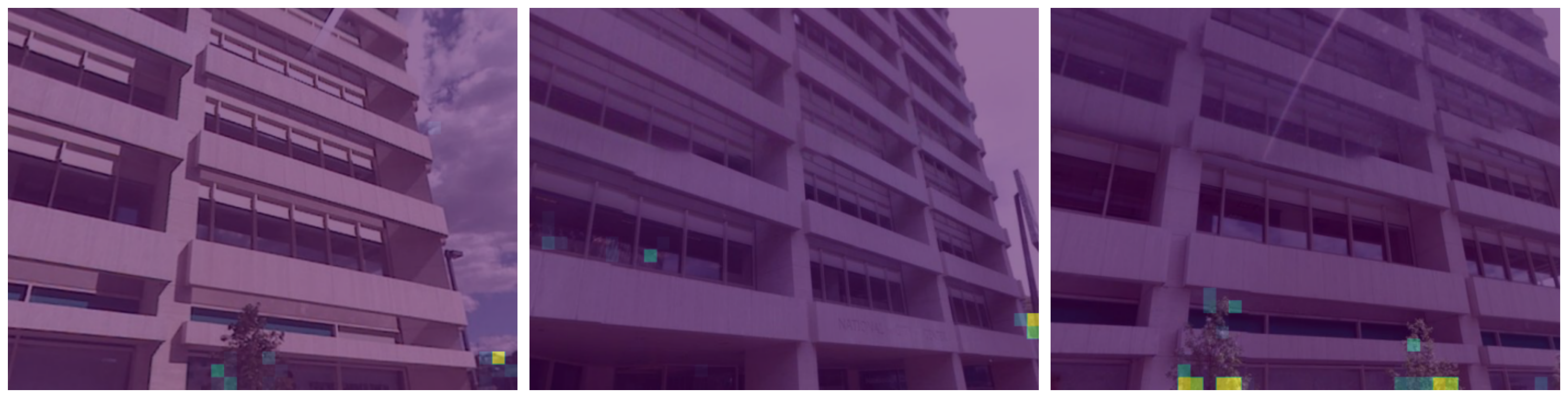}} 
     \\
     
     \rotatebox{90}{\hspace{0.5cm}SA (ours)} &

     \multicolumn{3}{c}{\includegraphics[width=\scaleone]{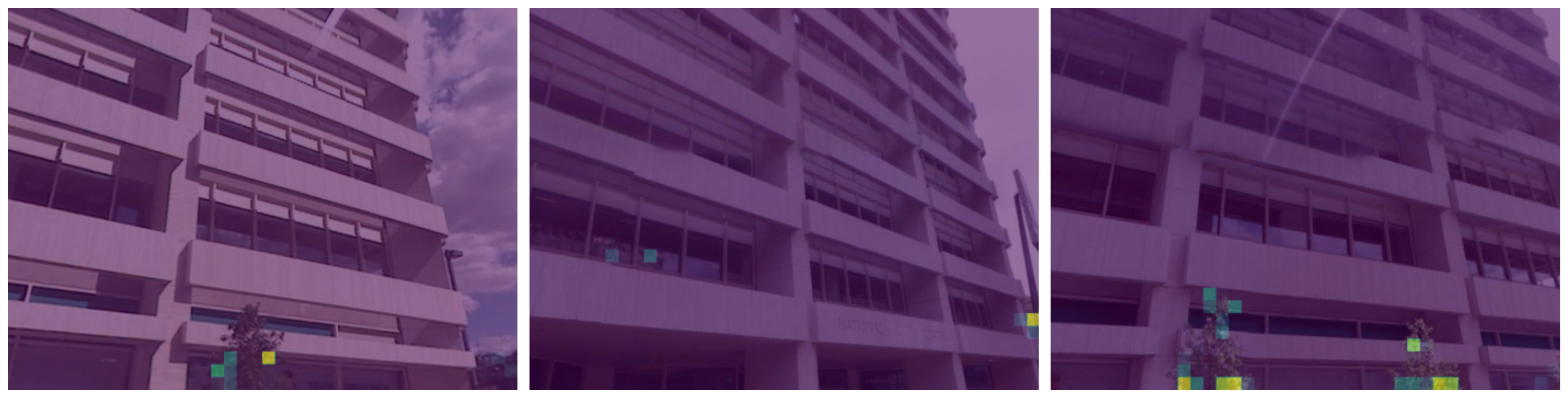}} 
     \\

     \rotatebox{90}{\hspace{0.5cm}Burstiness} &

     \multicolumn{3}{c}{\includegraphics[width=\scaleone]{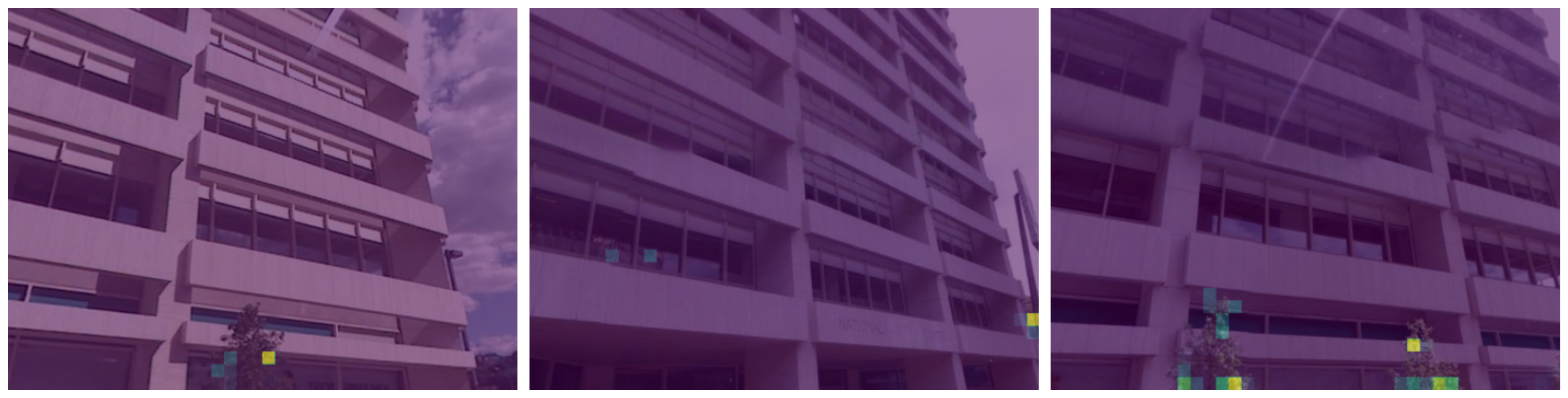}} 
     \\

    \end{tabular}
    \caption{In this example from the Pitts30k dataset, we highlight the improved behavior of local features, learnt through \ourM{} with burstiness weighting. Thus, we show the soft assignment weighting using both the vanilla NetVLAD model as well as the proposed \ourM{} model. The weighting patterns for the negative and the positive remain similar across the rows. However, in the query image, it can be observed that \ourM{}'s soft assignment selects the overlapping region between the tree and the building at the bottom center of the image, whereas NetVLAD's soft assigment selects the trees at the bottom right. The former is more consistent with the feature selection in the positive, thus improving the query-positive matching. Note that the burstiness weighting (4th row) had only a slight impact on top of the soft assignment weighting (3rd row), thus the variation in the weighting pattern between NetVLAD and \ourM{} is attributed more to feature-to-centroid distances than feature-to-feature distances in this case.}
    \label{fig:qual3}
\end{figure*}

\begin{figure*}
    \centering
    \begin{tabular}{
    p{0.15cm}
    >{\centering\arraybackslash}p{3cm}
    >{\centering\arraybackslash}p{3cm}
    >{\centering\arraybackslash}p{3cm}
    }
    & Query & Positive & Negative (Ours) 
    \\
    &
     \multicolumn{3}{c}{\includegraphics[width=\scaleone]{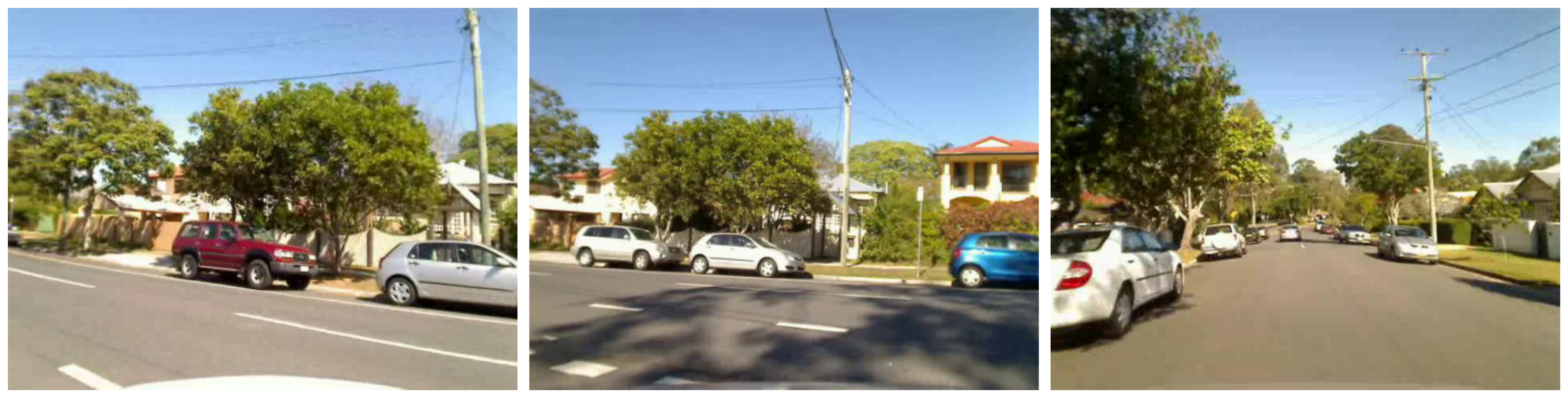}} 
     \\
     
     \rotatebox{90}{\hspace{0.5cm}SA (vanilla)} &

     \multicolumn{3}{c}{\includegraphics[width=\scaleone]{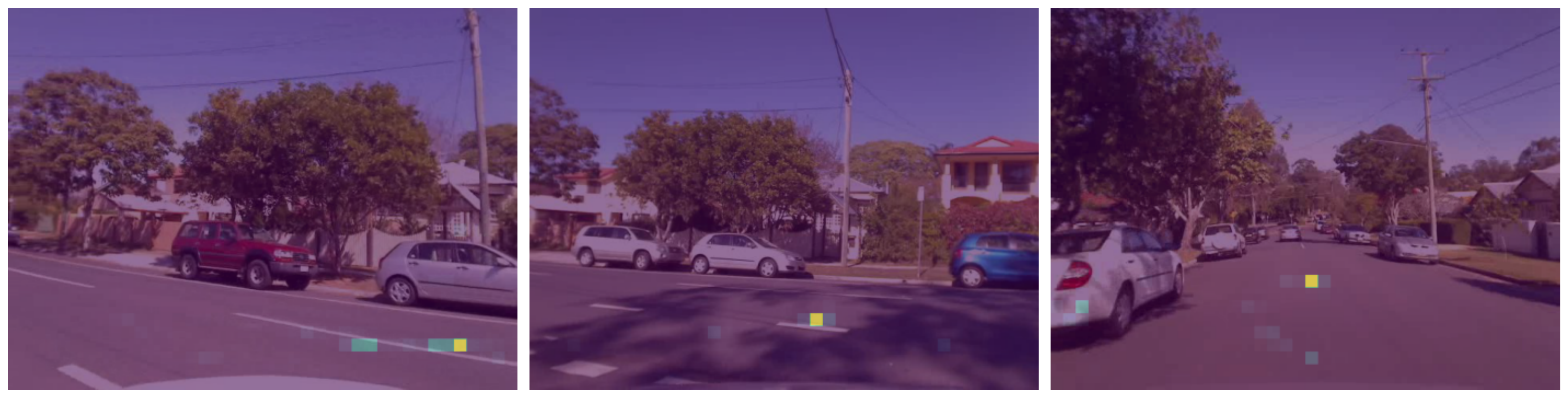}} 
     \\
     
     \rotatebox{90}{\hspace{0.5cm}SA (ours)} &

     \multicolumn{3}{c}{\includegraphics[width=\scaleone]{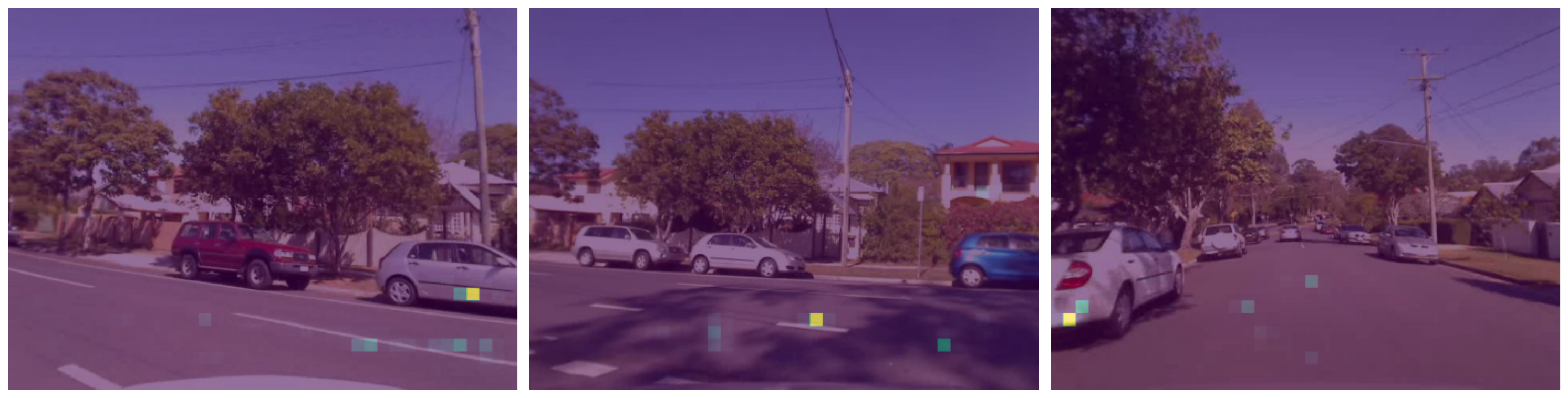}} 
     \\

     \rotatebox{90}{\hspace{0.5cm}Burstiness} &

     \multicolumn{3}{c}{\includegraphics[width=\scaleone]{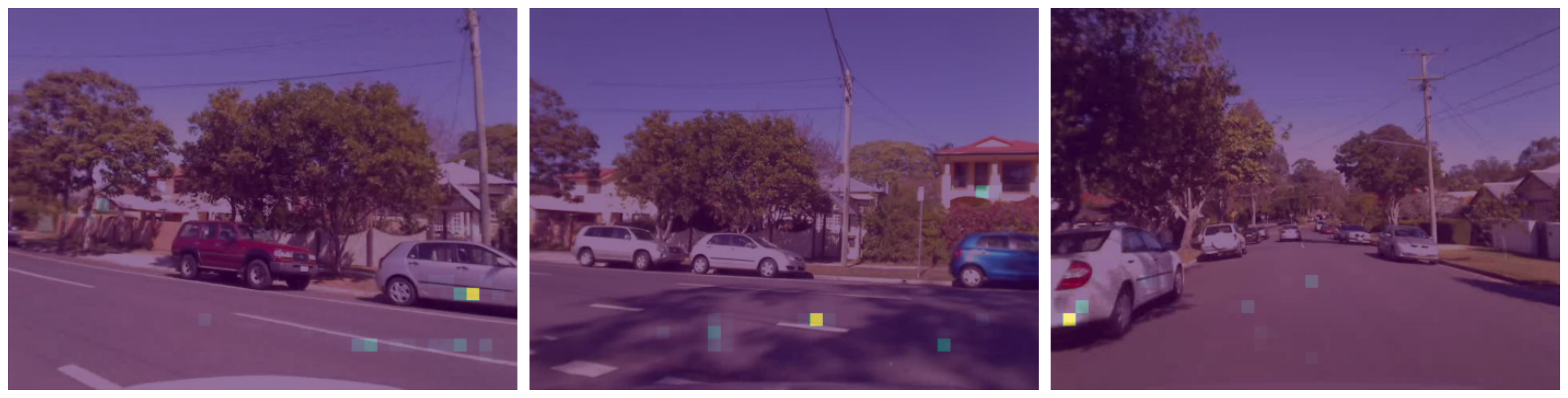}} 
     \\

    \end{tabular}
    \caption{In this example from the St Lucia dataset, we show a failure case of \ourM{}. Note that the middle column is the positive, correctly selected by the vanilla NetVLAD, and the third column is the negative, incorrectly selected by \ourM{}. Here, the soft assignment of \ourM{} model (3rd row) selects certain features on the car, consistently across the query (1st column) and the negative (3rd column), which leads to incorrect matching. On the other hand, NetVLAD's soft assignment (2nd row) selects features along the lane markings for both the query and the positive.}
    \label{fig:qual4}
\end{figure*}

\clearpage

\section*{Acknowledgements}
This work received funding from  ARC Laureate Fellowship FL210100156 to Michael Milford and the Centre for Augmented Reasoning at the Australian Institute for Machine Learning, University of Adelaide, Australia. The authors acknowledge continued support from the Queensland University of Technology (QUT) through the Centre for Robotics. S.H. also acknowledges continued support from the Embodied AI Cluster at CSIRO Robotics.
\bibliographystyle{splncs04}
\bibliography{main}

\begin{thebibliography}{10}
\providecommand{\url}[1]{\texttt{#1}}
\providecommand{\urlprefix}{URL }
\providecommand{\doi}[1]{https://doi.org/#1}

\bibitem{ali2022gsv}
Ali-bey, A., Chaib-draa, B., Gigu{\`e}re, P.: Gsv-cities: Toward appropriate supervised visual place recognition. Neurocomputing  \textbf{513},  194--203 (2022)

\bibitem{ali2023mixvpr}
Ali-bey, A., Chaib-draa, B., Gigu{\`e}re, P.: Mixvpr: Feature mixing for visual place recognition. In: Proceedings of the IEEE/CVF Winter Conference on Applications of Computer Vision. pp. 2998--3007 (2023)

\bibitem{arandjelovic2016netvlad}
Arandjelovic, R., Gronat, P., Torii, A., Pajdla, T., Sivic, J.: Netvlad: Cnn architecture for weakly supervised place recognition. In: Proceedings of the IEEE Conference on Computer Vision and Pattern Recognition. pp. 5297--5307 (2016)

\bibitem{arandjelovic2013all}
Arandjelovic, R., Zisserman, A.: All about vlad. In: Proceedings of the IEEE conference on Computer Vision and Pattern Recognition. pp. 1578--1585 (2013)

\bibitem{balntas2016learning}
Balntas, V., Riba, E., Ponsa, D., Mikolajczyk, K.: Learning local feature descriptors with triplets and shallow convolutional neural networks. In: Bmvc. vol.~1, p.~3 (2016)

\bibitem{berton2022rethinking}
Berton, G., Masone, C., Caputo, B.: Rethinking visual geo-localization for large-scale applications. In: Proceedings of the IEEE/CVF Conference on Computer Vision and Pattern Recognition. pp. 4878--4888 (2022)

\bibitem{berton2022deep}
Berton, G., Mereu, R., Trivigno, G., Masone, C., Csurka, G., Sattler, T., Caputo, B.: Deep visual geo-localization benchmark. In: Proceedings of the IEEE/CVF Conference on Computer Vision and Pattern Recognition. pp. 5396--5407 (2022)

\bibitem{berton2023eigenplaces}
Berton, G., Trivigno, G., Caputo, B., Masone, C.: Eigenplaces: Training viewpoint robust models for visual place recognition. In: Proceedings of the IEEE/CVF International Conference on Computer Vision (ICCV). pp. 11080--11090 (October 2023)

\bibitem{camara2019spatio}
Camara, L.G., P{\v{r}}eu{\v{c}}il, L.: Spatio-semantic convnet-based visual place recognition. In: Eur. Conf. Mobile Robot. (2019)

\bibitem{cao2020unifying}
Cao, B., Araujo, A., Sim, J.: Unifying deep local and global features for image search. In: European Conference on Computer Vision. pp. 726--743. Springer (2020)

\bibitem{chen2021learning}
Chen, J., Hu, H., Wu, H., Jiang, Y., Wang, C.: Learning the best pooling strategy for visual semantic embedding. In: Proceedings of the IEEE/CVF conference on computer vision and pattern recognition. pp. 15789--15798 (2021)

\bibitem{chen2017deep}
Chen, Z., Jacobson, A., S{\"u}nderhauf, N., Upcroft, B., Liu, L., Shen, C., Reid, I., Milford, M.: Deep learning features at scale for visual place recognition. In: Robotics and Automation (ICRA), 2017 IEEE International Conference on. pp. 3223--3230. IEEE (2017)

\bibitem{chen2014convolutional}
Chen, Z., Lam, O., Jacobson, A., Milford, M.: Convolutional neural network-based place recognition. In: Australasian Conference on Robotics and Automation. vol.~2, p.~4 (2014)

\bibitem{chen2018learning}
Chen, Z., Liu, L., Sa, I., Ge, Z., Chli, M.: Learning context flexible attention model for long-term visual place recognition. IEEE Robotics and Automation Letters  \textbf{3}(4),  4015--4022 (2018)

\bibitem{chen2017only}
Chen, Z., Maffra, F., Sa, I., Chli, M.: Only look once, mining distinctive landmarks from convnet for visual place recognition. In: Intelligent Robots and Systems (IROS), 2017 IEEE/RSJ International Conference on. pp. 9--16. IEEE (2017)

\bibitem{cummins2008fab}
Cummins, M., Newman, P.: Fab-map: Probabilistic localization and mapping in the space of appearance. The International Journal of Robotics Research  \textbf{27}(6),  647--665 (2008)

\bibitem{delhumeau2013revisiting}
Delhumeau, J., Gosselin, P.H., J{\'e}gou, H., P{\'e}rez, P.: Revisiting the vlad image representation. In: Proceedings of the 21st ACM international conference on Multimedia. pp. 653--656. ACM (2013)

\bibitem{dosovitskiy2021an}
Dosovitskiy, A., Beyer, L., Kolesnikov, A., Weissenborn, D., Zhai, X., Unterthiner, T., Dehghani, M., Minderer, M., Heigold, G., Gelly, S., Uszkoreit, J., Houlsby, N.: An image is worth 16x16 words: Transformers for image recognition at scale. In: International Conference on Learning Representations (2021), \url{https://openreview.net/forum?id=YicbFdNTTy}

\bibitem{garg2021where}
Garg, S., Fischer, T., Milford, M.: Where is your place, visual place recognition? In: Proceedings of the International Joint Conference on Artificial Intelligence, {IJCAI}. pp. 4416--4425 (8 2021)

\bibitem{garg2021seqnet}
Garg, S., Milford, M.J.: Seqnet: Learning descriptors for sequence-based hierarchical place recognition. IEEE Robotics and Automation Letters  (2021)

\bibitem{garg2024robohop}
Garg, S., Rana, K., Hosseinzadeh, M., Mares, L., Suenderhauf, N., Dayoub, F., Reid, I.: Robohop: Segment-based topological map representation for open-world visual navigation. In: 2024 IEEE International Conference on Robotics and Automation (ICRA) (2024)

\bibitem{garg2018lost}
Garg, S., Suenderhauf, N., Milford, M.: Lost? appearance-invariant place recognition for opposite viewpoints using visual semantics. In: Proceedings of Robotics: Science and Systems XIV (2018)

\bibitem{gawel2018x}
Gawel, A., Del~Don, C., Siegwart, R., Nieto, J., Cadena, C.: X-view: Graph-based semantic multi-view localization. IEEE Robotics and Automation Letters  \textbf{3}(3),  1687--1694 (2018)

\bibitem{guissous19}
Guissous, K., Gouet-Brunet, V.: Saliency and burstiness for feature selection in cbir. In: European Workshop on Visual Information Processing (EUVIP). pp. 111--116 (2019)

\bibitem{hausler2021patchnetvlad}
Hausler, S., Garg, S., Xu, M., Milford, M., Fischer, T.: Patch-netvlad: Multi-scale fusion of locally-global descriptors for place recognition. In: Proceedings of the IEEE/CVF Conference on Computer Vision and Pattern Recognition. pp. 14141--14152 (2021)

\bibitem{ho2007detecting}
Ho, K.L., Newman, P.: Detecting loop closure with scene sequences. International Journal of Computer Vision  \textbf{74}(3),  261--286 (2007)

\bibitem{hong2019TextPlace}
Hong, Z., Petillot, Y., Lane, D., Miao, Y., Wang, S.: Textplace: Visual place recognition and topological localization through reading scene texts. In: 2019 IEEE/CVF International Conference on Computer Vision (ICCV). pp. 2861--2870 (2019). \doi{10.1109/ICCV.2019.00295}

\bibitem{izquierdo2023optimal}
Izquierdo, S., Civera, J.: Optimal transport aggregation for visual place recognition (2023)

\bibitem{jegou2009burstiness}
J{\'e}gou, H., Douze, M., Schmid, C.: On the burstiness of visual elements. In: 2009 IEEE conference on computer vision and pattern recognition. pp. 1169--1176. IEEE (2009)

\bibitem{jegou2010aggregating}
J{\'e}gou, H., Douze, M., Schmid, C., P{\'e}rez, P.: Aggregating local descriptors into a compact image representation. In: IEEE Conf. Comput. Vis. Pattern Recog. pp. 3304--3311 (2010)

\bibitem{keetha2023anyloc}
Keetha, N., Mishra, A., Karhade, J., Jatavallabhula, K.M., Scherer, S., Krishna, M., Garg, S.: Anyloc: Towards universal visual place recognition. IEEE Robotics and Automation Letters  (2023)

\bibitem{keetha2021hierarchical}
Keetha, N.V., Milford, M., Garg, S.: A hierarchical dual model of environment-and place-specific utility for visual place recognition. IEEE Robotics and Automation Letters  \textbf{6}(4),  6969--6976 (2021)

\bibitem{khaliq2022multires}
Khaliq, A., Milford, M., Garg, S.: Multires-netvlad: Augmenting place recognition training with low-resolution imagery. IEEE Robotics and Automation Letters  \textbf{7}(2),  3882--3889 (2022)

\bibitem{kmiec2018learnable}
Kmiec, S., Bae, J., An, R.: Learnable pooling methods for video classification. arXiv preprint arXiv:1810.00530  (2018)

\bibitem{le2020city}
Le, D.C., Youn, C.H.: City-scale visual place recognition with deep local features based on multi-scale ordered {VLAD} pooling. arXiv preprint arXiv:2009.09255  (2020)

\bibitem{linnextvlad18}
Lin, R., Xiao, J., Fan, J.: Nextvlad: An efficient neural network to aggregate frame-level features for large-scale video classification. In: European Conference on Computer Vision Workshops (2018)

\bibitem{lowry2016visual}
Lowry, S., S{\"u}nderhauf, N., Newman, P., Leonard, J.J., Cox, D., Corke, P., Milford, M.J.: Visual place recognition: A survey. IEEE Transactions on Robotics  \textbf{32}(1),  1--19 (2016)

\bibitem{manandharbursty}
Manandhar, D., Yap, K.H.: Feature repetitiveness similarity metrics in visual search. IEEE Signal Processing Letters pp. 1368--1372 (2017)

\bibitem{masone2021survey}
Masone, C., Caputo, B.: A survey on deep visual place recognition. IEEE Access  \textbf{9},  19516--19547 (2021)

\bibitem{mereu2022learning}
Mereu, R., Trivigno, G., Berton, G., Masone, C., Caputo, B.: Learning sequential descriptors for sequence-based visual place recognition. IEEE Robotics and Automation Letters  \textbf{7}(4),  10383--10390 (2022)

\bibitem{miech2017learnable}
Miech, A., Laptev, I., Sivic, J.: Learnable pooling with context gating for video classification. In: IEEE Conf. Comput. Vis. Pattern Recog. Worksh. (2017)

\bibitem{milford2008mapping}
Milford, M., Wyeth, G.: Mapping a suburb with a single camera using a biologically inspired slam system. IEEE Transactions on Robotics  \textbf{24}(5),  1038--1053 (2008)

\bibitem{milford2012seqslam}
Milford, M.J., Wyeth, G.F.: Seqslam: Visual route-based navigation for sunny summer days and stormy winter nights. In: Robotics and Automation (ICRA), 2012 IEEE International Conference on. pp. 1643--1649. IEEE (2012)

\bibitem{neubert2019neurologically}
Neubert, P., Schubert, S., Protzel, P.: A neurologically inspired sequence processing model for mobile robot place recognition. IEEE Robotics and Automation Letters  \textbf{4}(4),  3200--3207 (2019)

\bibitem{nowicki2017real}
Nowicki, M.R., Wietrzykowski, J., Skrzypczy{\'n}ski, P.: Real-time visual place recognition for personal localization on a mobile device. Wireless Personal Communications  \textbf{97},  213--244 (2017)

\bibitem{oertel2020augmenting}
Oertel, A., Cieslewski, T., Scaramuzza, D.: Augmenting visual place recognition with structural cues. arXiv preprint arXiv:2003.00278  (2020)

\bibitem{oquab2023dinov2}
Oquab, M., Darcet, T., Moutakanni, T., Vo, H.V., Szafraniec, M., Khalidov, V., Fernandez, P., Haziza, D., Massa, F., El-Nouby, A., Howes, R., Huang, P.Y., Xu, H., Sharma, V., Li, S.W., Galuba, W., Rabbat, M., Assran, M., Ballas, N., Synnaeve, G., Misra, I., Jegou, H., Mairal, J., Labatut, P., Joulin, A., Bojanowski, P.: Dinov2: Learning robust visual features without supervision (2023)

\bibitem{paolicelli2022learning}
Paolicelli, V., Tavera, A., Masone, C., Berton, G., Caputo, B.: Learning semantics for visual place recognition through multi-scale attention. In: Image Analysis and Processing--ICIAP 2022: 21st International Conference, Lecce, Italy, May 23--27, 2022, Proceedings, Part II. pp. 454--466. Springer (2022)

\bibitem{pengattention21}
Peng, G., Zhang, J., Li, H., Wang, D.: Attentional pyramid pooling of salient visual residuals for place recognition. In: IEEE/CVF International Conference on Computer Vision (ICCV) (2021)

\bibitem{radenovic2018fine}
Radenovi{\'c}, F., Tolias, G., Chum, O.: Fine-tuning cnn image retrieval with no human annotation. IEEE Trans. Pattern Anal. Mach. Intell.  \textbf{41}(7),  1655--1668 (2018)

\bibitem{revaud2019learning}
Revaud, J., Almaz{\'a}n, J., Rezende, R.S., Souza, C.R.d.: Learning with average precision: Training image retrieval with a listwise loss. In: Proceedings of the IEEE International Conference on Computer Vision. pp. 5107--5116 (2019)

\bibitem{saltontext}
Salton, G., Buckley, C.: Term-weighting approaches in automatic text retrieval. Information Processing and Management pp. 513--523 (1988)

\bibitem{sandler2018mobilenetv2}
Sandler, M., Howard, A., Zhu, M., Zhmoginov, A., Chen, L.C.: Mobilenetv2: Inverted residuals and linear bottlenecks. In: Proceedings of the IEEE conference on computer vision and pattern recognition. pp. 4510--4520 (2018)

\bibitem{schubert2023visual}
Schubert, S., Neubert, P., Garg, S., Milford, M., Fischer, T.: Visual place recognition: A tutorial. RAM  (2023)

\bibitem{schubert2021fast}
Schubert, S., Neubert, P., Protzel, P.: Fast and memory efficient graph optimization via icm for visual place recognition. In: Robotics: Science and Systems (2021)

\bibitem{shah2023gnm}
Shah, D., Sridhar, A., Bhorkar, A., Hirose, N., Levine, S.: Gnm: A general navigation model to drive any robot. In: 2023 IEEE International Conference on Robotics and Automation (ICRA). pp. 7226--7233. IEEE (2023)

\bibitem{shi2015early}
Shi, M., Avrithis, Y., J{\'e}gou, H.: Early burst detection for memory-efficient image retrieval. In: Proceedings of the IEEE Conference on Computer Vision and Pattern Recognition. pp. 605--613 (2015)

\bibitem{sivic2003video}
Sivic, J., Zisserman, A.: Video google: A text retrieval approach to object matching in videos. In: Proceedings of International Conference on Computer Vision (ICCV). p.~1470. IEEE (2003)

\bibitem{sun2020dagc}
Sun, Q., Liu, H., He, J., Fan, Z., Du, X.: Dagc: Employing dual attention and graph convolution for point cloud based place recognition. In: Proceedings of the 2020 International Conference on Multimedia Retrieval. pp. 224--232 (2020)

\bibitem{sunderhauf2013we}
S{\"u}nderhauf, N., Neubert, P., Protzel, P.: Are we there yet? challenging seqslam on a 3000 km journey across all four seasons. In: Proc. of Workshop on Long-Term Autonomy, IEEE International Conference on Robotics and Automation (ICRA). p.~2013 (2013)

\bibitem{suomela2023placenav}
Suomela, L., Kalliola, J., Dag, A., Edelman, H., K{\"a}m{\"a}r{\"a}inen, J.K.: Placenav: Topological navigation through place recognition. In: 2024 IEEE International Conference on Robotics and Automation (ICRA) (2024)

\bibitem{thoma2020soft}
Thoma, J., Paudel, D.P., Van~Gool, L.: Soft contrastive learning for visual localization. Advances in Neural Information Processing Systems 33  (2020)

\bibitem{tolias2015particular}
Tolias, G., Sicre, R., J{\'e}gou, H.: Particular object retrieval with integral max-pooling of cnn activations. In: International Conference on Learning Representations (2016)

\bibitem{torii201524}
Torii, A., Arandjelovic, R., Sivic, J., Okutomi, M., Pajdla, T.: 24/7 place recognition by view synthesis. In: Proceedings of the IEEE Conference on Computer Vision and Pattern Recognition. pp. 1808--1817 (2015)

\bibitem{toriivprrepet}
Torii, A., Sivic, J., Pajdla, T., Okutomi, M.: Visual place recognition with repetitive structures. In: Proceedings of the IEEE/CVF Conference on Computer Vision and Pattern Recognition. pp. 883--890 (2013)

\bibitem{trichetbursty19}
Trichet, R., O'Connor, N.E.: Gaussian normalization: Handling burstiness in visual data. In: IEEE International Conference on Advanced Video and Signal Based Surveillance (AVSS). pp.~1--8 (2019)

\bibitem{tsintotas2022revisiting}
Tsintotas, K.A., Bampis, L., Gasteratos, A.: The revisiting problem in simultaneous localization and mapping: A survey on visual loop closure detection. IEEE Transactions on Intelligent Transportation Systems  \textbf{23}(11),  19929--19953 (2022)

\bibitem{uy2018pointnetvlad}
Uy, M.A., Lee, G.H.: Pointnetvlad: Deep point cloud based retrieval for large-scale place recognition. In: Proceedings of the IEEE conference on computer vision and pattern recognition. pp. 4470--4479 (2018)

\bibitem{wang2018cosface}
Wang, H., Wang, Y., Zhou, Z., Ji, X., Gong, D., Zhou, J., Li, Z., Liu, W.: Cosface: Large margin cosine loss for deep face recognition. In: Proceedings of the IEEE conference on computer vision and pattern recognition. pp. 5265--5274 (2018)

\bibitem{wang2022transvpr}
Wang, R., Shen, Y., Zuo, W., Zhou, S., Zheng, N.: Transvpr: Transformer-based place recognition with multi-level attention aggregation. In: Proceedings of the IEEE/CVF Conference on Computer Vision and Pattern Recognition. pp. 13648--13657 (2022)

\bibitem{wang2019multi}
Wang, X., Han, X., Huang, W., Dong, D., Scott, M.R.: Multi-similarity loss with general pair weighting for deep metric learning. In: Proceedings of the IEEE Conference on Computer Vision and Pattern Recognition. pp. 5022--5030 (2019)

\bibitem{warburg2020mapillary}
Warburg, F., Hauberg, S., L{\'o}pez-Antequera, M., Gargallo, P., Kuang, Y., Civera, J.: Mapillary street-level sequences: A dataset for lifelong place recognition. In: Proceedings of the IEEE/CVF Conference on Computer Vision and Pattern Recognition. pp. 2626--2635 (2020)

\bibitem{xia2021soe}
Xia, Y., Xu, Y., Li, S., Wang, R., Du, J., Cremers, D., Stilla, U.: Soe-net: A self-attention and orientation encoding network for point cloud based place recognition. In: Proceedings of the IEEE/CVF Conference on Computer Vision and Pattern Recognition (CVPR) (2021)

\bibitem{xin2019real}
Xin, Z., Cui, X., Zhang, J., Yang, Y., Wang, Y.: Real-time visual place recognition based on analyzing distribution of multi-scale cnn landmarks. J. Intell. Robot. Syst.  \textbf{94}(3-4),  777--792 (2019)

\bibitem{yildiz2022amstertime}
Yildiz, B., Khademi, S., Siebes, R.M., van Gemert, J.: Amstertime: A visual place recognition benchmark dataset for severe domain shift. arXiv preprint arXiv:2203.16291  (2022)

\bibitem{yin2022general}
Yin, P., Zhao, S., Cisneros, I., Abuduweili, A., Huang, G., Milford, M., Liu, C., Choset, H., Scherer, S.: General place recognition survey: Towards the real-world autonomy age. arXiv preprint arXiv:2209.04497  (2022)

\bibitem{yu2019spatial}
Yu, J., Zhu, C., Zhang, J., Huang, Q., Tao, D.: Spatial pyramid-enhanced netvlad with weighted triplet loss for place recognition. IEEE Trans. Neural Netw. Learn. Syst.  \textbf{31}(2),  661--674 (2019)

\bibitem{zaffar2021vpr}
Zaffar, M., Garg, S., Milford, M., Kooij, J., Flynn, D., McDonald-Maier, K., Ehsan, S.: Vpr-bench: An open-source visual place recognition evaluation framework with quantifiable viewpoint and appearance change. International Journal of Computer Vision pp. 1--39 (2021)

\bibitem{zhangpcan19}
Zhang, W., Xiao, C.: Pcan: 3d attention map learning using contextual information for point cloud based retrieval. In: IEEE/CVF Conference on Computer Vision and Pattern Recognition (CVPR). pp. 12428--12437 (2019)

\bibitem{zhangsurvey}
Zhang, X., Wang, L., Su, Y.: Visual place recognition: A survey from deep learning perspective. Pattern Recognition  \textbf{113},  107760 (2021)

\bibitem{zhong2019ghostvlad}
Zhong, Y., Arandjelovi{\'c}, R., Zisserman, A.: Ghostvlad for set-based face recognition. In: Computer Vision--ACCV 2018: 14th Asian Conference on Computer Vision, Perth, Australia, December 2--6, 2018, Revised Selected Papers, Part II 14. pp. 35--50. Springer (2019)

\bibitem{zhou2016learning}
Zhou, B., Khosla, A., Lapedriza, A., Oliva, A., Torralba, A.: Learning deep features for discriminative localization. In: Proceedings of the IEEE Conference on Computer Vision and Pattern Recognition. pp. 2921--2929 (2016)

\bibitem{zhu2018attention}
Zhu, Y., Wang, J., Xie, L., Zheng, L.: Attention-based pyramid aggregation network for visual place recognition. In: 2018 ACM Multimedia Conference on Multimedia Conference. pp. 99--107. ACM (2018)

\end{thebibliography}
\end{document}